\documentclass{article}
\usepackage[preprint]{neurips_2025}

\usepackage{amsmath}
\usepackage{amsfonts}
\usepackage{amssymb}
\usepackage{bbm}
\usepackage{nicefrac}

\usepackage{graphicx}
\usepackage{wrapfig}
\usepackage{tikz}

\usepackage{booktabs}
\usepackage{multirow}
\usepackage{makecell}

\usepackage[table]{xcolor}
\usepackage{colortbl}

\usepackage{microtype}
\usepackage{nicefrac}
\usepackage{enumitem}

\usepackage[svgnames]{xcolor}
\definecolor{eccvblue}{rgb}{0.12,0.49,0.85}
\usepackage[pagebackref,breaklinks,colorlinks,citecolor=eccvblue]{hyperref}

\usepackage{algorithmic}
\usepackage[ruled]{algorithm2e}

\usepackage[utf8]{inputenc}
\usepackage[T1]{fontenc}
\usepackage{bbding}
\usepackage[skins]{tcolorbox}
\usepackage{titletoc}
\usepackage{capt-of}
\usepackage{caption}

\definecolor{cvprblue}{HTML}{DAE8F5} 
\definecolor{cvprgray}{HTML}{EAEAEA}
\usepackage[accsupp]{axessibility}

\DeclareRobustCommand\onedot{\futurelet\@let@token\@onedot}
\def\@onedot{\ifx\@let@token.\else.\null\fi\xspace}

\def\eg{\emph{e.g}\onedot}

\usepackage{makecell}

\usepackage{capt-of}
\usepackage{caption}
\usepackage{graphicx}

\newtcolorbox{findingbox}{
    enhanced,
    colback=gray!8,          
    colframe=cvprblue,  
    arc=3pt,                 
    boxrule=1pt,           
    drop shadow=gray!50,     
    left=4pt, right=4pt, top=2pt, bottom=2pt, 
    fontupper=\small 
}

\title{AIBench: Evaluating Visual-Logical Consistency in Academic Illustration Generation}

\author{Zhaohe~Liao$^{1,2*}$, Kaixun~Jiang$^{3*}$, Zhihang~Liu$^{1,4*}$, Yujie~Wei$^{3*}$, Junqiu~Yu$^{1, 3*}$, \\ \textbf{Quanhao~Li$^{1, 3*}$}, 
\textbf{Hong-Tao~Yu$^{1, 5*}$}, \textbf{Pandeng~Li$^{1\dagger}$, Yuzheng~Wang$^{1}$, Zhen~Xing$^{1}$,} \\ \textbf{Shiwei~Zhang$^{1}$, Chen-Wei~Xie$^{1}$, Yun~Zheng$^{1}$, Xihui~Liu$^{6\dagger}$}\\
    {$^{1}$Tongyi Lab, Alibaba Group} \quad {$^{2}$SJTU} \quad {$^{3}$FDU} \quad {$^{4}$USTC}
    \quad {$^{5}$SEU} \quad {$^{6}$HKU} \\\\
    Project page: \url{https://deep-kaixun.github.io/aibench-page/}
    \vspace{-3mm}
}

\begin{document}
\maketitle
\begingroup
\renewcommand{\thefootnote}{}
\footnotetext{* Equal contribution, random order. $^{\dagger}$ Corresponding authors.}
\endgroup
\setcounter{footnote}{0}

\begin{abstract}
  Although image generation has boosted various applications via its rapid evolution, whether the state-of-the-art models are able to produce ready-to-use academic illustrations for papers is still largely unexplored.
  Directly comparing or evaluating the illustration with VLM is native but requires oracle multi-modal understanding ability, which is unreliable for long and complex texts and illustrations.
  To address this, we propose AIBench, the first benchmark using VQA for evaluating logic correctness of the academic illustrations and VLMs for assessing aesthetics.
  In detail, we designed four levels of questions proposed from a logic diagram summarized from the method part of the paper, which query whether the generated illustration aligns with the paper on different scales.
  Our VQA-based approach raises more accurate and detailed evaluations on visual-logical consistency while relying less on the ability of the judger VLM.
  With our high-quality AIBench, we conduct extensive experiments and conclude that the performance gap between models on this task is significantly larger than general ones, reflecting their various complex reasoning and high-density generation ability. 
  Further, the logic and aesthetics are hard to optimize simultaneously as in handcrafted illustrations.
  Additional experiments further state that test-time scaling on both abilities significantly boosts the performance on this task.
\end{abstract}
\section{Introduction}
\label{sec:intro}

\begin{figure*}[t]
    \centering
    \includegraphics[width=1.0\linewidth]{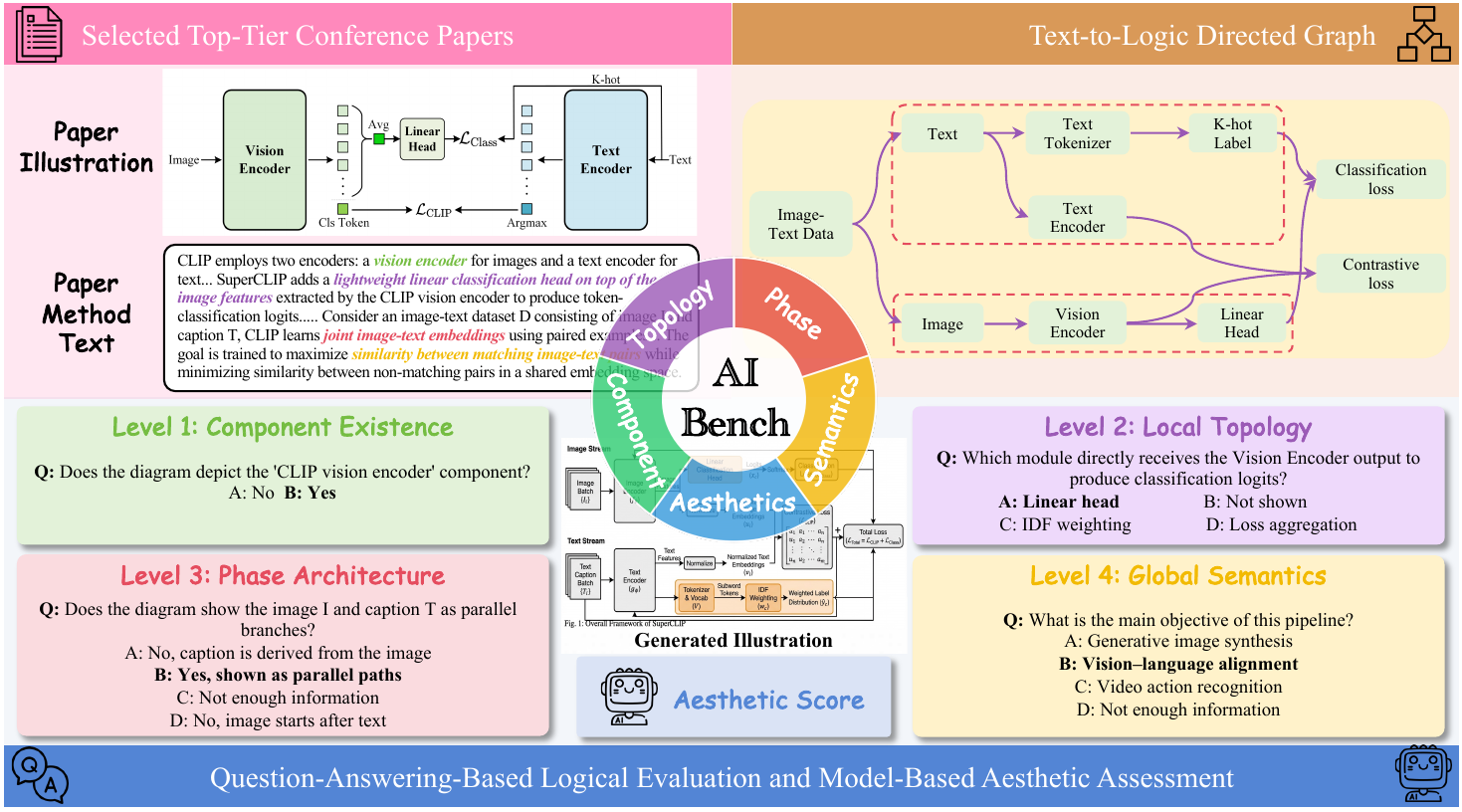}
    \caption{\textbf{Overview of AIBench.} We introduce a comprehensive benchmark to evaluate academic illustration generation through two primary dimensions: question-answering-based logical evaluation and model-based aesthetic assessment. To systematically assess logical accuracy, we collect top-tier conference papers, construct text-to-logic directed graphs, and manually annotate QA pairs across four hierarchical levels.
    Ultimately, our AIBench establishes a new evaluation standard for generated academic illustrations.
    }
    \label{fig:teaser}
\end{figure*}

Image generation has experienced significant growth in recent years~\cite{sd, sd3, dalle3, flux, t2i_r1}.
Such advances~\cite{januspro, showo, emu3, blip3o, bagel, uniworld, omnigen2, qwenimage,wan26,seedance15pro,kline-omni} have enabled significantly broader applications~\cite{tuo2024anytext2, liu2024glyph, chen2024textdiffuser, qwenimage,autoposter, postermaker, dreamposter, wei2025dreamrelation, showtable, wei2025routing,jiang2026genagent,jiang2025enhancing}, easing the workload of graphic design.
However, when focusing on academic scenarios, generating an academic illustration (\emph{i.e.}, the framework of a method) still heavily relies on handcrafting.
Since this task faces a unique challenge, the ability of state-of-the-art (SOTA) models~\cite{wan26,seedance15pro,kline-omni,qwenimage,team2025zimage} on this task remains largely unexplored.
Specifically, as the model needs to take the method part of the paper as input and target generating the framework figure representing its core contribution, it needs to understand the details of the paper and conclude its main contribution, while being able to present such details and contributions in both an aesthetic and logically correct manner.
Moreover, judging whether such a generated image aligns with the logic in the paper could be challenging for vision language models (VLMs).

Some works~\cite{zhu2026paperbanana,zhu2026autofigure} have made an initial attempt to evaluate such ability. However, they hide the aforementioned challenge by directly using VLMs to compare figures as a whole, assuming the oracle ability of VLMs, which could be suspicious when the underlying logic of the paper is non-trivial, and the generated figures are logically complex.
Moreover, the influence of components in the illustration remains unclear, leading to an uninterpretable assessment.

Focusing on these challenges, instead of using VLMs to compare illustrations as a whole, we propose a \textbf{visual question answering} (VQA) style logic assessment and model style aesthetic assessment.
Specifically, for logic assessment, we model the assessment as a sequence of VQA tasks, where the questions query the generated illustration at multiple levels to ensure the questions cover all necessary points of the underlying logic of the paper.
Specifically, the questions are proposed according to four different levels from low-level components to high-level semantics, namely component-existence-level, local-topology-level, phase-architecture-level, and global-semantics-level.
It is worth noting that the crucial text generation ability is implicitly embedded in our QAs, as answering the questions requires correctly generated characters in the illustration.
An example of these questions is provided in~Fig.~\ref{fig:teaser}.
To generate high-quality QAs for these levels, we first acquire Gemini 3 Flash~\cite{gemini3} to generate a logic graph from the method text, then propose the questions from the generated logic graph at each corresponding level with proper prompts.
A more detailed introduction to the framework is in Sec.~\ref{sec:method_pipeline}.
With this framework, we are able to comprehensively query whether the generated illustration aligns with the original paper.
As for aesthetic assessment, such an objective-based evaluation suits VLMs better. 
Overall, a higher accuracy reflects a higher quality of the generated illustration, and the accuracy at different levels provides a more detailed evaluation.
A comparison between our benchmark and existing ones is in Table~\ref{tab:benchmark-compare}.

Following this philosophy, we propose \textbf{AIBench}, a high-quality benchmark for evaluating the \textbf{A}cademic \textbf{I}llustration generation ability of models.
It contains 300 open-access papers accepted by top conferences with 5704 corresponding QA pairs for checking the necessary properties that an ideal illustration should have.
The QAs are automatically generated via the aforementioned framework and carefully checked by multiple human experts to ensure their high quality.
A more detailed dataset distribution is presented in~Fig.~\ref{fig:data1}.

We conduct extensive experiments on our benchmark and conclude valuable findings.
First, we conclude that the gap between models on the academic illustration generation task is more profound than that on general tasks, indicating that the performance difference between models is still significant when facing long and complex text reasoning for understanding and high-density content in generation.
Moreover, we conclude that aesthetics and logic are somewhat of a trade-off, which also exists in handcrafted illustrations, as more text descriptions and complex layouts may lead to better logic, while breaking aesthetics and vice versa.
We further point out that both strong reasoning ability and high-density generation ability are required for this academic illustration task, and test-time scaling strategies on both sides can shatter the capability ceilings of current models.
More detailed discussions are in~Sec.~\ref{sec:expr}. In conclusion, our contributions can be summarized as follows:
\begin{enumerate}
\item We propose AIBench, the first VQA-based benchmark for evaluating the academic illustration generation ability of models, providing high-quality QAs for evaluating both logic consistency and aesthetics.
\item We design a scalable QA proposing framework that raises high-quality QAs from a summarized logic diagram to evaluate the academic illustration generation ability from low-level content to high-level semantics.
\item We conduct experiments on SOTA open- and closed-source unified models and T2I models along with test-time scaling methods on our benchmark to give a full view of current capabilities on academic illustration generation.
\end{enumerate}

\begin{table}[t]
    \centering
    \caption{\textbf{Comparison between our AIBench and related benchmarks.}}
    \resizebox{\textwidth}{!}{%
    \setlength\tabcolsep{4pt}
    \begin{tabular}{lccccc}
    \toprule
         \textbf{Benchmark} & \textbf{Data Construction} & \textbf{\# Papers} & \textbf{\# Avg. Eval Unit} & \textbf{Eval Method} & \textbf{Granularity} \\
    \midrule
         PaperBanana~\cite{zhu2026paperbanana} & Auto & 292 & 1 & VLM-as-Judge & Coarse \\
         AutoFigure~\cite{zhu2026autofigure} & Auto & 3,300 & 1 & VLM-as-Judge & Coarse \\
         \textbf{AIBench (Ours)} & Auto + Human & 300 & 19.01 & QA-Based & Fine-grained \\
    \bottomrule
    \end{tabular}
    }
    \label{tab:benchmark-compare}
    \vspace{-5pt}
\end{table}
\section{Related Work}

\noindent\textbf{Text-To-Image Generation Benchmarks.}
Along with rapid progress in text-to-image generation~\cite{qwenimage,wan26,omnigen2,chen2025blip3onext,team2025zimage,seedance15pro,flux}, diverse benchmarks have been proposed to evaluate complementary aspects of model capability. GenEval~\cite{ghosh2023geneval} primarily tests compositional prompt following, WISE~\cite{niu2025wise} targets knowledge-intensive and factual consistency, and T2I-CompBench++~\cite{huang2025t2i} provides a broader suite for compositional reasoning with tailored automatic metrics. Beyond general-purpose T2I evaluation, recent benchmarks move toward academic figure generation: PaperBananaBench~\cite{zhu2026paperbanana} evaluates methodology diagrams with VLM-as-Judge and reference-based scoring, while FigureBench~\cite{zhu2026autofigure} targets long-context scientific illustration with VLM-based referenced scoring and pairwise comparison. Unlike VLM-judge protocols prone to evaluator instability, our AIBench builds logic-grounded multiple-choice QA pairs from paper content, enabling stable, reproducible assessment of content completeness, logical coherence, communicative clarity, and visual quality.

\noindent\textbf{Automated Scientific Figure Generation.}
With the progress of generative models, there is growing interest in automating scientific figure creation~\cite{qiang2016learning,PPSGen,xu2022posterbot}. Early work mainly follows \emph{extract-and-layout} pipelines that rearrange existing paper content (\textit{e.g.}, PPSGen~\cite{PPSGen} focuses on text-only slides). Recent systems move toward \emph{agentic, code-based rendering} for more controllable outputs, such as PPTAGENT~\cite{PPTAGENT} and Paper2Poster~\cite{pang2025paper2poster}; however, they still largely reorganize and stylize source assets rather than generating figures from scratch.
Despite these advances, recent work has begun to generate scientific figures from scratch directly from paper content. PaperBanana~\cite{zhu2026paperbanana} introduces an end-to-end agentic pipeline for producing academic illustrations; however, it typically conditions on limited inputs (\textit{e.g.}, method excerpts and captions), which can encourage style imitation while missing fine-grained technical details. AutoFigure~\cite{zhu2026autofigure} proposes a reasoned rendering paradigm that converts long-form text into a machine-readable symbolic blueprint, but at the cost of a heavier multi-stage pipeline with increased system complexity and deployment overhead.

\section{AIBench}
\begin{figure*}[t]
    \centering \includegraphics[width=0.99\linewidth]{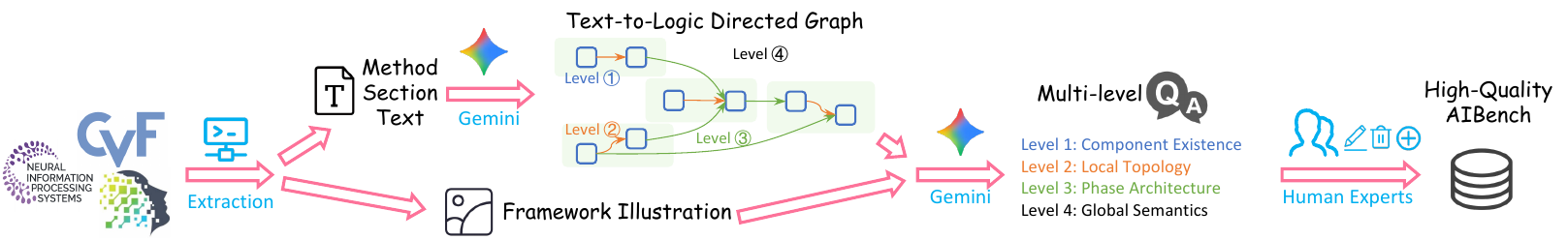}
    \caption{\textbf{QA data construction pipeline of AIBench.} We construct the multi-level QA pairs with the help of Gemini, followed by careful human annotation.}
    \label{fig:method_pipeline}
\end{figure*}

\noindent\textbf{Motivation.} Evaluating generated academic illustrations requires a comprehensive consideration of both logical accuracy and aesthetic quality. Existing benchmarks~\cite{zhu2026autofigure,zhu2026paperbanana} predominantly adopt a VLM-as-Judge paradigm, employing a single VLM to simultaneously score logic and aesthetics. This introduces ``metric ambiguity'' by conflating objective logical errors with subjective aesthetic flaws, resulting in a lack of fine-grained interpretability. To this end, our AIBench explicitly decouples the evaluation process into two distinct dimensions: objective logic and subjective aesthetics. The logical assessment focuses on elements like core components, data flows, and text rendering, which can be objectively cross-checked against the method descriptions from multiple perspectives. Specifically, we formulate logical evaluation into a fine-grained Visual Question Answering (VQA) task using multiple-choice QA pairs. To ensure data quality, we design an automated multi-level QA construction pipeline equipped with human-in-the-loop annotation (Fig.~\ref{fig:method_pipeline}). Meanwhile, the subjective aesthetic assessment is independently evaluated via a meticulously selected aesthetic model to reliably simulate human preferences. The following sections detail Data Curation, QA Construction, Data Annotation, Data Analysis, and Evaluation Protocol.

\subsection{Data Collection \& Curation}
We source our benchmark from papers in 2025 across four premier AI conferences (CVPR, ICCV, ICLR, NeurIPS).
Using recent, arXiv-available papers mitigates data contamination (these papers are less likely to have been incorporated into the training corpora) and provides LaTeX sources for precise text-figure alignment. Our pipeline extracts methodology text and primary architecture figures in three steps: \textit{(i)} \textbf{Text extraction}: LaTeX files are parsed to isolate the methodology section, aided by an LLM fallback. \textit{(ii)} \textbf{Figure extraction}: Candidate figures are ranked via rule-based heuristics on filenames and captions, followed by a VLM-driven pairwise selection for the top-$k$. \textit{(iii)} \textbf{Quality verification}: A VLM confirms the paired figure is a valid pipeline diagram. Finally, uniform sampling across conferences yields approximately 3,000 diverse text-figure pairs.

\subsection{QA Construction}\label{sec:method_pipeline}
We adopt a two-stage pipeline: \textit{(i)} constructing a structured logic-directed graph from the method text, and \textit{(ii)} generating multi-level QA pairs based on this graph and an academic framework illustration from the original paper.

\noindent\textbf{Stage 1: Text-to-Logic Directed Graph Construction.}
Methodology sections in academic papers are often lengthy and weakly structured, making direct QA generation from raw text prone to logical inconsistencies. We therefore introduce a structured intermediate representation. Since AI methods are typically described as modular pipelines or network architectures with explicit data flows, their computation naturally maps to a directed graph, which serves as an effective intermediate form. Motivated by this, we use Gemini 3 Flash to parse the methodology text and construct a logic-directed graph
$\mathcal{G}=(\mathcal{V},\mathcal{E},\mathcal{P})$, where $\mathcal{V}$ denotes key components or data artifacts,
$\mathcal{E}$ denotes directed data flows, and $\mathcal{P}$ denotes architectural phases (\textit{e.g.}, ``pre-training'').
To ensure faithfulness, we constrain the model to preserve the original terminology from the source paper. This intermediate representation converts unstructured text into an explicit structure, facilitating subsequent QA synthesis.

\noindent\textbf{Stage 2: Multi-Level QA Generation.}
Based on node-, edge-, phase-, and graph-level attributes of $\mathcal{G}$, we generate QA pairs at four levels:
component, topology, phase, and semantics.
We deploy four level-specific QA generators (Gemini 3 Flash), each focusing on the corresponding structural aspects of $\mathcal{G}$.
To align the question abstraction with human diagram conventions, we additionally provide the original illustration as a reference image.
The four levels are defined as follows:

$\quad\bullet$ \textbf{Level 1: Component Existence.}
This level verifies the presence and completeness of key nodes $v \in \mathcal{V}$ by checking whether core components specified in the directed graph also appear in the reference image (and identifying critical missing elements). Since these queries require recognizing labels and basic glyphs, they also probe text render accuracy.

$\quad\bullet$ \textbf{Level 2: Local Topology.}
This level examines local connectivity and direct transfers between adjacent nodes via edges $e \in \mathcal{E}$, testing whether outputs from upstream components are correctly routed to the intended downstream modules. As answering such questions depends on tracing lines and arrowheads, it also reflects the visual clarity of local layout organization.

$\quad\bullet$ \textbf{Level 3: Phase Architecture.}
Leveraging phase annotations $\mathcal{P}$, this level targets macro-architectural organization across phases, including parallel branches, feature aggregation (macro fan-in), branching (macro fan-out), and global feedback loops. Because these patterns rely on cross-module spatial composition, the questions also indicate the coherence of  overall layout.

$\quad\bullet$ \textbf{Level 4: Global Semantics.}
By combining the method text with the whole $\mathcal{G}$, this level formulates questions about the system’s end-to-end design intent and task paradigm. Since such high-level understanding requires integrating evidence across the entire diagram, performance at this level also reflects the figure’s global visual clarity.

In conclusion, by proceeding from the intrinsic logical hierarchy of the figure, we design this pipeline in a bottom-up manner, ensuring it effectively covers all aspects of the verifiable dimensions for evaluating academic illustrations.

\subsection{Data Filtering \& Annotation}
\label{sec:data_filter}

To guarantee high quality and accuracy of our benchmark, we design a filtering pipeline that integrates automated screening with rigorous human inspection~\cite{liu2025capability}.

\noindent\textbf{Step 1: Accuracy-based Screening.}
We generate QA pairs following the pipeline in Sec.~\ref{sec:method_pipeline}, then instruct Gemini 3 Flash to answer them based strictly on the original figures. Samples with low overall QA accuracy are discarded as their QA pairs are likely inaccurate.

\noindent\textbf{Step 2: Hallucination-based Filtering.}
For incorrectly answered questions, we additionally provide Gemini 3.1 Pro with the original method text. If the model still fails to answer correctly given the ground-truth context, the question is deemed a hallucination artifact and removed.
These two automated steps effectively eliminate the vast majority of inaccurate or unanswerable QA pairs.

\noindent\textbf{Step 3: Human Expert Review.}
Human experts systematically review all surviving QA pairs for factual correctness, reasoning difficulty, and completeness.
Simplistic questions are revised for greater complexity, and new challenging questions are added to ensure comprehensiveness, ultimately yielding a curated dataset of \textbf{300} samples with 5704 high-quality QA pairs.

\subsection{Data Analysis}
\begin{figure}[t]
    \centering
    \includegraphics[width=\linewidth]{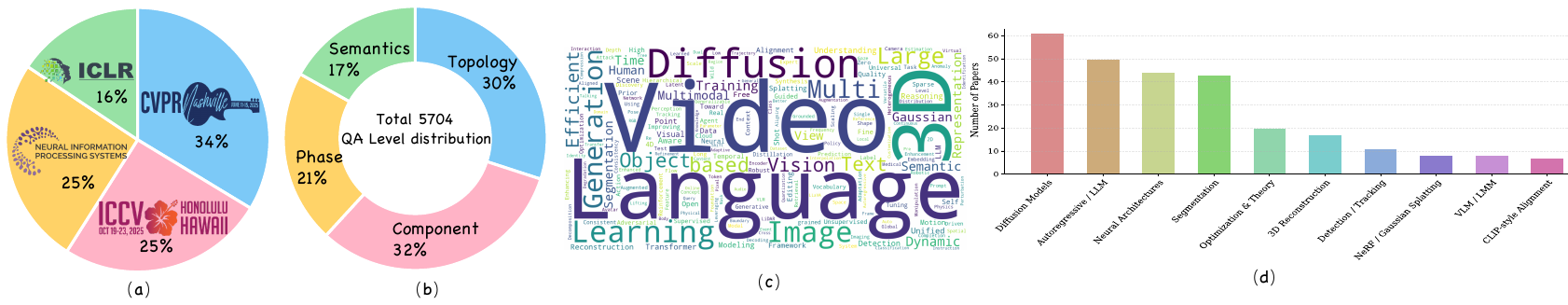}
    \caption{ \textbf{Statistics analysis of AIBench.}  (a) Papers are curated from four representative 2025 conferences (CVPR, ICCV, NeurIPS, ICLR). (b) Models are evaluated on four hierarchical QA levels: Component, Topology, Phase, and Semantics. (c) Word cloud shows lexical frequency and topic diversity of our AIBench. (d) Top research topics in 2025 research papers, covering diffusion, LLMs, 3D reconstruction, etc.}
    \label{fig:data1}
\end{figure}

\noindent\textbf{Data Source.} Our benchmark is constructed from papers published in 2025 at four premier conferences: CVPR (34\%), ICCV (25\%), NeurIPS (25\%), and ICLR (16\%). This selection ensures high technical quality through rigorous peer review and maintains topical freshness. By drawing from both CV and ML communities, the dataset provides balanced coverage of vision, generative modeling, and multimodal learning, effectively avoiding low-quality data source.

\noindent\textbf{Topic Distribution.} As illustrated in Fig~\ref{fig:data1}(a), the benchmark covers major research directions in year 2025, including diffusion models, language modeling, and multimodal reasoning. While diffusion and LLMs represent a substantial portion—reflecting current research trends, other emerging fields are consistently represented to prevent single-paradigm dominance. This diverse composition requires models to demonstrate both generative capability and structured understanding across heterogeneous semantic settings.

\noindent\textbf{QA Hierarchy.} To evaluate the consistency between framework diagrams and text, we designed 5,704 QA pairs across four hierarchical levels (Fig~\ref{fig:data1}(d)). The Component (32\%) and Topology (30\%) levels verify fine-grained nodes and local connectivities. Conversely, the Phase (21\%) and Semantic (17\%) levels assess macro-architectural organization and global design alignment. This bottom-up structure ensures a precise, multi-granularity assessment.

\begin{figure*}[t]
    \centering
    \includegraphics[width=1.0\linewidth]{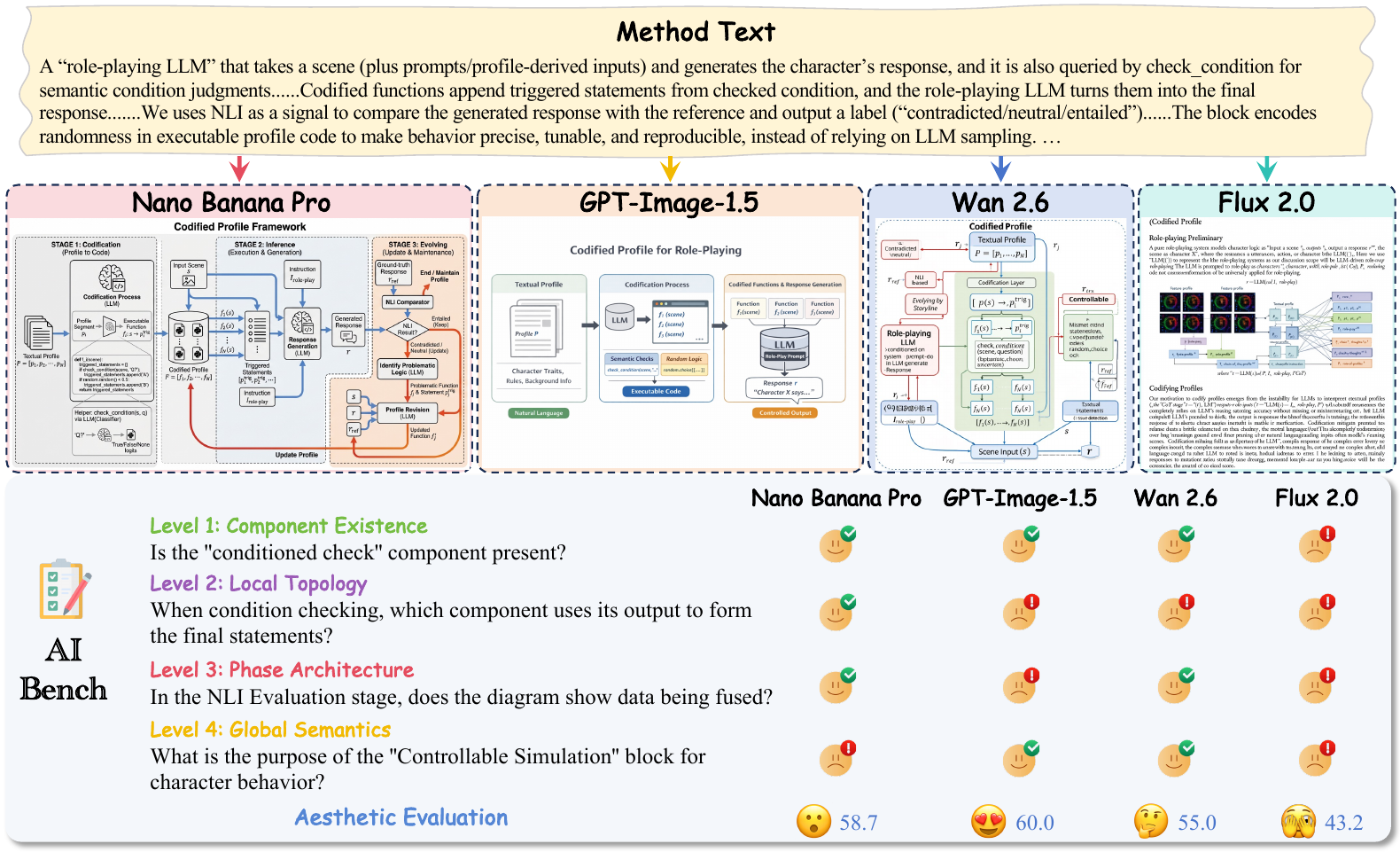}
    \caption{\textbf{The evaluation pipeline of AIBench.} Models generate academic illustrations based on method descriptions, and are subsequently evaluated on two primary dimensions: logical accuracy, assessed via paper-specific QA pairs across four hierarchical levels (L1–L4, ranging from component existence to global semantics), and visual appeal, measured by a model-based aesthetic score.}
    \label{fig:intro_pipeline}
\end{figure*}

\subsection{Evaluation Protocol}

Our evaluation framework is decoupled into two complementary parts. First, we formulate the objective logic assessment as a Visual Question Answering (VQA) task based on our constructed QA pairs, explicitly covering the objective dimensions of academic illustrations. Second, for the subjective aesthetic evaluation, we employ a model-based evaluation to simulate human visual preferences. Figure~\ref{fig:intro_pipeline} illustrates the complete evaluation pipeline.

\noindent\textbf{Objective Logic Evaluation.}
The logic evaluation is formalized as a visual question answering (VQA) task. For each test sample, let $V$ denote the generated image provided by the user (conditioned on the original method text), and $\mathcal{Q}$ be the corresponding set of multi-level questions constructed in Section~\ref{sec:method_pipeline}. Given a multi-modal model solver $\mathcal{F}$, the evaluation process is defined as deriving the predicted answer $\hat{a} = \mathcal{F}(V, q)$ for each question $q \in \mathcal{Q}$. Notably, the solver strictly relies on the visual input $V$ without any supplementary textual context. By decomposing the complex global logic into atomic QA pairs, our framework fundamentally reduces the reasoning burden on $\mathcal{F}$. We adopt Qwen3-VL-235B-A22B-Instruct~\cite{bai2025qwen3} as the default solver. Figure~\ref{fig:robust}(a) shows the robustness of our evaluation across different solvers.

Specifically, logic accuracy is computed independently for each of the four hierarchical levels. Let $\mathcal{Q}_l = \bigcup_{i=1}^{N} \mathcal{Q}_l^{(i)}$ denote the complete set of level-$l$ questions across all $N$ samples, where $\mathcal{Q}_l^{(i)}$ is the subset of questions derived from the $i$-th sample at level $l$. The accuracy at level $l$ is defined as:
\begin{equation}
    \text{Acc}_l = \frac{1}{|\mathcal{Q}_l|} \sum_{q \in \mathcal{Q}_l} \mathbbm{1}[ \hat{a_q} = a_q^*],
\end{equation}
where $\hat{a_q}$ and $a_q^*$ denote the solver's predicted option and the ground-truth answer for question $q$, respectively. Crucially, we adopt question-level global averaging rather than sample-level averaging. This design ensures that each visual reasoning judgment contributes equally to the final metric, naturally yielding more stable statistical estimates for architecturally complex illustrations that inherently generate more questions.

\noindent\textbf{Subjective Aesthetic Evaluation.}
\label{sec:aesthetic_pipeline}
Beyond logical accuracy, the aesthetic quality of academic illustration is also crucial for an academic paper.
Currently, aesthetic evaluation in image generation relies on CLIP-based metrics (\textit{e.g.,} Aesthetic Score~\cite{aesthetic_score}, PickScore~\cite{kirstain2023pick}), VLM-as-Judge paradigms (\textit{e.g.,} AutoFig~\cite{zhu2026autofigure}), or specialized aesthetic models (\textit{e.g.,} UniPercept~\cite{cao2025unipercept}, AesR1~\cite{aesR1}). Since no existing metric is tailored for the unique graphical characteristics of academic illustrations, we conduct a systematic preliminary study to identify the most reliable evaluator by using all 300 generated academic illustrations from our AIBench.

The quantitative results in Table~\ref{tab:aesthetic_comparison} reveal that traditional aesthetic metrics struggle with the academic illustration generation task. CLIP-based metrics demonstrate severe misalignment and insensitivity; for instance, PickScore assigns its highest value to the human-ranked worst model, while Aesthetic Score exhibits negligible variance. 
General-purpose VLMs (\textit{e.g.,} Qwen3-VL-235b-a22b-instruct~\cite{bai2025qwen3} evaluated in Table~\ref{tab:aesthetic_comparison}) struggle to reliably assess aesthetic quality, particularly for academic illustrations, due to a lack of aesthetic knowledge. In contrast, specialized aesthetic models, even without domain-specific fine-tuning on academic figures, are capable of evaluating their aesthetic quality, aligning more closely with human preferences.
Based on these observations, we adopt UniPercept as our definitive aesthetic metric, as it offers superior evaluation efficiency. Built upon InternVL3-8B~\cite{zhu2025internvl3}, UniPercept evaluates images across complementary perceptual domains, outputting continuous ratings from 0 to 100. In AIBench, we utilize its Image Aesthetics score to evaluate the holistic visual appeal, composition, and harmonic unity of the generated academic illustrations.

\begin{table}[t]
\centering
\caption{\textbf{Comparison of different aesthetic evaluation metrics} and human ranking on generated academic illustrations.}
\label{tab:aesthetic_comparison}
\setlength{\tabcolsep}{4pt}
\resizebox{0.95\textwidth}{!}{
\begin{tabular}{lcccccc}
\toprule
\textbf{Method} & 
\makecell{\textbf{Aesthetic} \\ \textbf{Score}} & 
\makecell{\textbf{Pick} \\ \textbf{Score}} & 
\makecell{\textbf{VLM-as} \\ \textbf{-Judge}} &
\textbf{UniPercept} & 
\textbf{AesR1} & 
\makecell{\textbf{Human} \\ \textbf{Rank}} \\
\midrule
Original Figure & \underline{5.193} & 19.423 & \textbf{8.92} & 51.11 & 62.39 & 3 \\
Nano Banana Pro~\cite{nano_banana} & 5.181 & \underline{19.904} & 8.63 & \underline{55.04} & \textbf{64.82} & \underline{2} \\
GPT-Image-1.5~\cite{gpt-image-1.5} & \textbf{5.210} & 19.717 & \underline{8.87} & \textbf{57.50} & \underline{64.71} & \textbf{1} \\
Wan 2.6~\cite{wan26} & 5.151 & 19.866 & 7.60 & 51.50 & 59.09 & 4 \\
FLUX2-dev~\cite{flux} & 5.150 & \textbf{20.127} & 5.53 & 42.40 & 46.48 & 5 \\
\bottomrule
\end{tabular}
}
\end{table}

\vspace{1mm}
\vspace{1mm}
\noindent\textbf{Overall Score.} To ensure numerical consistency, we scale the logic accuracy ($\text{Acc}_l$) to a 0--100 range to match the aesthetic metric. Our evaluation comprises four objective logic dimensions plus one subjective aesthetic dimension, and the Overall Score is computed as the arithmetic mean of all five. By combining verifiable QA with human-aligned aesthetic judgment, this dual-track protocol provides a comprehensive, standardized, and interpretable benchmark for academic illustration generation.
\section{Experiments}
\label{sec:expr}
\subsection{Experiment Settings}
To provide a comprehensive analysis of the current landscape in academic illustration generation, we evaluate a diverse suite of state-of-the-art multi-modal generation models, including \textit{Closed-Source Models}: Seedream 4.5~\cite{seedream2025seedream}, Seedream 5.0~\cite{seedream-5-0-lite}, Wan 2.6~\cite{wan26}, GPT-Image-1.5~\cite{gpt-image-1.5}, and Nano Banana Pro~\cite{nano_banana}; \textit{Open-Source T2I Models}: Qwen-Image~\cite{qwenimage}, Qwen-Image-2512~\cite{qwenimage}, Z-Image~\cite{team2025zimage}, and FLUX2-dev~\cite{flux}; \textit{Open-Source Unified Models}: BAGEL~\cite{bagel}, UniWorld-V1~\cite{uniworld}, OmniGen2~\cite{omnigen2}, BLIP3o-NEXT~\cite{chen2025blip3onext}, and Emu3.5~\cite{cui2025emu3}.
We employ Qwen3-VL-235B-A22B-Instruct~\cite{bai2025qwen3} as the VQA solver for all generated images, and UniPercept~\cite{cao2025unipercept} for the aesthetic evaluation. See more detailed settings in the Appendix.

\begin{table*}[!t]
\centering
\caption{\textbf{The evaluation results on AIBench} of different state-of-the-art models. We report level-wise and overall performance across our defined five dimensions.}
\label{tab:model_comparison}
\setlength{\tabcolsep}{4pt} 

\renewcommand{\arraystretch}{1.2} 
\resizebox{0.94\textwidth}{!}{%
\begin{tabular}{l c c c c c c >{\columncolor{cvprblue}}c}
\toprule
\textbf{Methods} & \textbf{Size} & \textbf{Component} & \textbf{Topology} & \textbf{Phase} & \textbf{Semantics} & \textbf{Aesthetics} & \textbf{Score} \\ \midrule
Original Image & - & 82.65 & 57.98 & 79.57 & 79.13 & 51.11 & 70.09 \\ \midrule
\multicolumn{7}{l}{\textit{\textcolor{gray}{{{Closed-Source Models}}}}} \\
Seedream 4.5~\cite{seedream2025seedream} & - & 67.89 & 52.42 & 48.14 & 74.47 & 55.48 & 59.68 \\
Wan 2.6~\cite{wan26} & - & 68.60 & 56.11 & 72.56 & 80.43 & 51.50 & 65.84 \\
Seedream 5.0~\cite{seedream-5-0-lite} & - & 82.93 & 72.81 & 72.10 & 86.53 & 51.78 & 73.23 \\
GPT-Image-1.5~\cite{gpt-image-1.5} & - & 66.23 & 50.87 & 55.95 & 77.55 & \textbf{57.50} & 61.62 \\
Nano Banana Pro~\cite{nano_banana} & - & \textbf{87.80} & \textbf{74.81} & \textbf{82.67} & \textbf{88.54} & 55.04 & \textbf{77.77} \\ \midrule
\multicolumn{7}{l}{\textcolor{gray}{\textit{{{Open-Source T2I Models}}}}} \\
Qwen-Image~\cite{qwenimage} & 27B & 21.92 & 24.31 & 44.81 & 42.16 & 32.69 & 33.18 \\
Qwen-Image-2512~\cite{qwenimage} & 27B & \textbf{32.27} & 29.11 & 39.95 & \textbf{56.39} & \textbf{56.45} & \textbf{42.83} \\
Z-Image~\cite{team2025zimage} & 6B &  25.61 & \textbf{34.42} & \textbf{57.41} & 54.00 & 36.65 & 41.62 \\
FLUX2-dev~\cite{flux} & 32B &  23.40 & 24.84 & 52.55 & 46.52 & 42.40 & 37.94 \\ \midrule
\multicolumn{7}{l}{\textcolor{gray}{\textit{{{Open-Source Unified Models}}}}} \\
BAGEL~\cite{bagel} & 7B &  1.16 & 10.34 & 27.43 & 3.12 & 37.21 & 15.85 \\
UniWorld-V1~\cite{uniworld} & 12B &  3.08 & 21.39 & 31.47 & 3.73 & 32.89 & 18.51 \\
OmniGen2~\cite{omnigen2} & 7B &  2.37 & 4.03 & 14.09 & 6.43 & 36.06 & 12.60 \\
BLIP3o-NEXT~\cite{chen2025blip3onext} & 3B &  0.50 & 3.04 & 2.22 & 0.31 & 47.21 & 10.66 \\
Emu3.5~\cite{cui2025emu3} & 34B & \textbf{31.04} & \textbf{29.90} & \textbf{43.49} & \textbf{57.78} & \textbf{48.53} & \textbf{42.15} \\ \bottomrule
\end{tabular}%
}
\end{table*}

\subsection{Main Results on AIBench}\label{subsec:main-results-aibench}

\noindent\textbf{Overall Performance and Model Hierarchy.} 
Table~\ref{tab:model_comparison} presents a comprehensive evaluation of various state-of-the-art models on AIBench. A prominent observation is the substantial performance gap between closed-source and open-source models. While existing general generative benchmarks (\textit{e.g.}, GenEval~\cite{ghosh2023geneval}) often present performance saturation, where the gap between open-source and closed-source models appears marginal, our AIBench reveals a starkly different reality, which demands high information density and deep logical reasoning over long texts. Closed-source models like Nano Banana Pro maintain robust performance (77.77), demonstrating a superior capacity for deep semantic understanding and dense information rendering. In contrast, open-source models struggle significantly, highlighting that in highly challenging framework diagram scenarios, closed-source models still hold an absolute advantage in complex reasoning and structural generation.

\begin{findingbox}
    \textit{Findings 1:} The performance saturation on general generative benchmarks masks a profound capability gap between open- and closed-source models when confronting the high-density, complex reasoning required for academic illustration generation.
\end{findingbox}

\begin{figure*}[!t]
    \centering
    \includegraphics[width=1.0\linewidth]{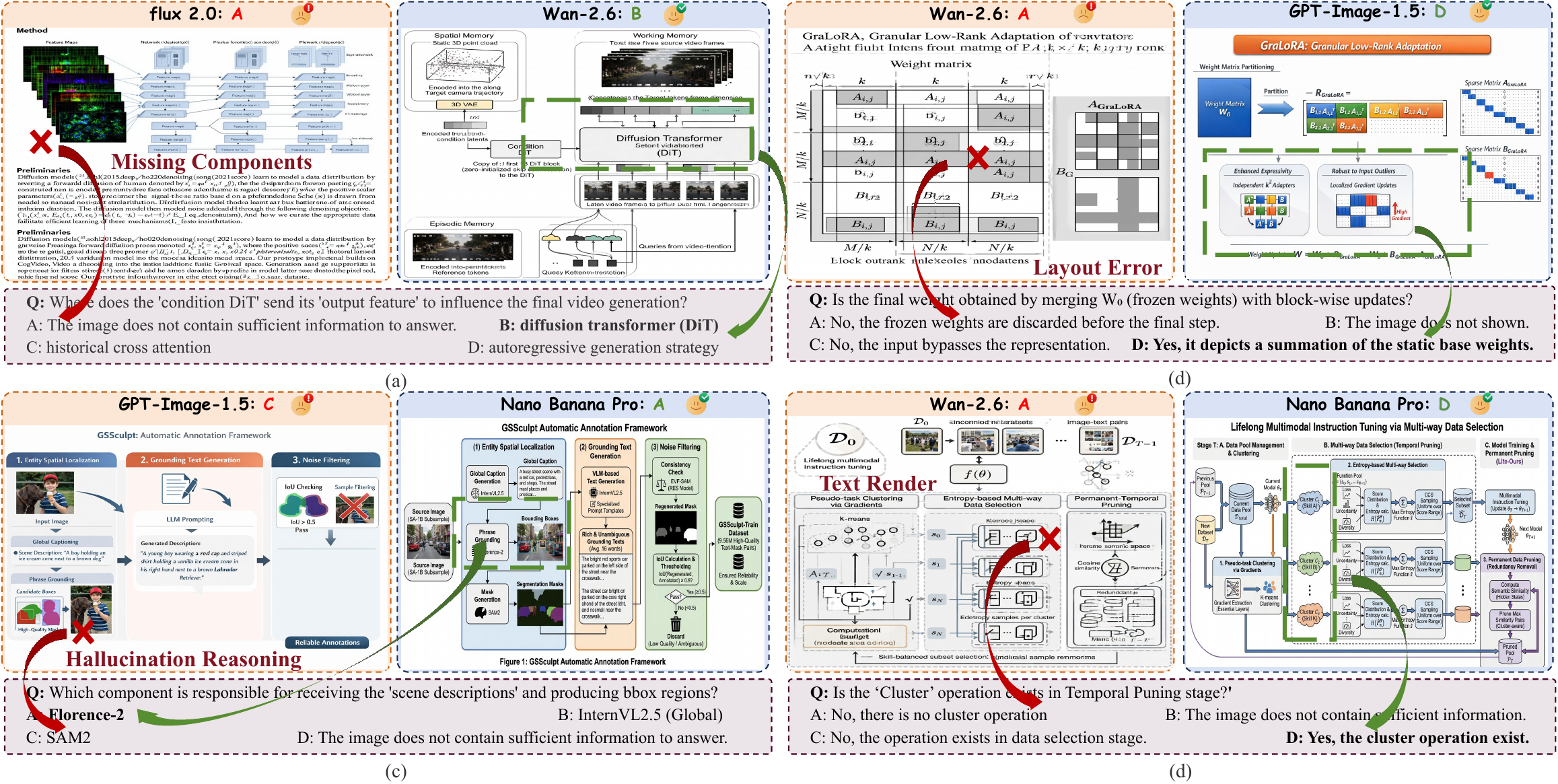}
    \caption{\textbf{Qualitative examples of typical generation failure modes} leading to incorrect answers: (a) missing components, (b) layout errors, (c) hallucinated reasoning/incorrect logic, and (d) unclear text rendering.}
    \label{fig:failure_cases}
\end{figure*}

\noindent\textbf{The Trade-off Between Logical Fidelity and Aesthetics.}
In the generation of complex, information-rich academic illustrations, a trade-off exists in academic illustration generation: models may struggle to optimize both rigorous logical constraints and aesthetic quality. Our evaluation demonstrates that generated results carrying high-density structural information typically achieve higher logical fidelity scores but suffer noticeable drops in aesthetic appeal. For example, GPT-Image-1.5 achieves the highest aesthetic quality among all closed-source models, yet its performance in logical dimensions remains uncompetitive. As shown in Fig.~\ref{fig:failure_cases}, the top-performing Nano Banana Pro model exhibits high information density with acceptable aesthetics. However, it still falls short of GPT; due to its lower information density, GPT achieves a cleaner and more visually appealing layout.

\begin{findingbox}
    \textit{Findings 2:} Models face a conflict mirroring human design challenges: logical completeness and visual aesthetics are often inherently exclusive and hard to balance.
\end{findingbox}

\noindent\textbf{Comparison with Original Images.}
Intriguingly, our evaluation reveals that the top-performing generated diagrams can surpass the \textit{Original Image} baseline (70.09) in terms of the overall score. Models like Nano Banana Pro (77.77) and Seedream 5.0 (73.23) outperform human-drawn original figures primarily in Component completeness and Global Semantic alignment. This counterintuitive result may mainly come from the subjectivity of humans. \textbf{Human authors often omit certain components, relying on implicit domain conventions}, while advanced models can produce diagrams that explicitly and exhaustively align with the methodology text.

\noindent\textbf{Typical Failure Modes in Logical Rendering.} As illustrated in Fig.~\ref{fig:failure_cases}, our qualitative analysis identifies four primary failure modes that hinder current models when translating complex texts into visual layouts. First, models like FLUX 2.0 frequently suffer from \textbf{missing components}, entirely omitting critical functional pathways (Fig.~\ref{fig:failure_cases}a). Second, even when components are present, models may exhibit \textbf{layout errors}; for instance, Wan-2.6 fails to spatially arrange elements to reflect the underlying mathematical summation (Fig.~\ref{fig:failure_cases}b). Third, \textbf{hallucinated reasoning} is prevalent, with models like GPT-Image-1.5 fabricating disconnected data flows or misattributing functions (Fig.~\ref{fig:failure_cases}c). Finally, poor \textbf{text rendering} and phase confusion (\textit{e.g.}, Wan-2.6 in Fig.~\ref{fig:failure_cases}d) severely degrade logical usability by obscuring architectural stages. In contrast, models with superior ability (\textit{e.g.}, Nano Banana Pro) effectively mitigate these issues by accurately grounding textual logic into structural visual layouts. These failures underscore that academic illustration generation is a rigorous visual reasoning challenge rather than merely an aesthetic pixel arrangement for most models.

\begin{table*}[!t]
\centering
\caption{Exploration on the different test time scaling methods to increase the performance on AIBench. We use different models based on their distinct bottlenecks.}
\label{tab:method_comparison}
\setlength{\tabcolsep}{4pt} 
\renewcommand{\arraystretch}{1.2} 
\resizebox{0.94\textwidth}{!}{%
\begin{tabular}{l c c c c c >{\columncolor{cvprblue}}c}
\toprule
\textbf{Methods} & \textbf{Component} & \textbf{Topology} & \textbf{Phase} & \textbf{Semantics} & \textbf{Aesthetics} & \textbf{Overall} \\ \midrule
\multicolumn{6}{l}{\textcolor{gray}{\textit{{{Rewriting}}}}} \\
Qwen-Image-2512~\cite{qwenimage} & 32.27 & 29.11 & 39.95 & 56.39 & 56.45 & 42.83 \\
Rewritten Qwen-Image-2512 & \textbf{56.97} & \textbf{45.93} & \textbf{57.71} & \textbf{71.97} & \textbf{59.35} & \textbf{58.39} \\ \midrule
\multicolumn{6}{l}{\textcolor{gray}{\textit{{{AutoFigure Pipeline~\cite{zhu2026autofigure}}}}}} \\
SVG Code (Gemini-2.5-Flash~\cite{gemini2.5pro}) & 87.51 & \textbf{79.66} & 81.90 & 91.98 & 43.12 & 76.83 \\
Nano Banana Pro~\cite{nano_banana}  & \textbf{87.80} & 74.81 & \textbf{82.67} & 88.54 & 55.04 & 77.77 \\ 
SVG Prompted Nano Banana Pro & 87.57 & 76.50 & 78.91 & \textbf{92.14} & \textbf{55.05} & \textbf{78.03} \\ \midrule
\multicolumn{6}{l}{\textcolor{gray}{\textit{{{Post Enhancement}}}}} \\
Wan2.6~\cite{wan26} & 68.60 & 56.11 & 72.56 & 80.43 & 51.50 & 65.84 \\
Wan2.6 \textit{w/} BoN & 69.90 & 58.50 & \textbf{74.24} & 80.53 & 52.97 & 67.23 \\
Wan2.6 \textit{w/} Edit & \textbf{75.70} & \textbf{61.80} & 72.81 & \textbf{82.02} & \textbf{54.21} & \textbf{69.31} \\ \bottomrule
\end{tabular}%
}
\end{table*}

\subsection{Roadmap Towards Logical Visual Generation.}
To further push the boundaries of academic illustration generation, we explore the potential of Test-Time Scaling (TTS). 
The inherent difficulty of this task lies in its dual requirements: it demands not only deep comprehension and structural planning over long, complex methodology texts but also highly robust visual rendering capabilities.
Consequently, we demonstrate that applying TTS to either the reasoning or generation aspect can effectively boost overall performance.
We apply targeted enhancements to different models based on their distinct bottlenecks. The quantitative results of these explorations are detailed in Table~\ref{tab:method_comparison}.

\noindent\textbf{Scaling the Reasoning Phase.}
Different from MLLMs~\cite{liu2025hybrid}, Open-source T2I models typically struggle with the long-context comprehension required to parse raw methodology texts. To mitigate this, we introduce a \textit{Rewriting} strategy~\cite{showtable}, utilizing a powerful LLM Qwen-Max to act as a planner. The LLM pre-reasons over the method text and condenses it into a highly structured prompt. As shown in Table~\ref{tab:method_comparison}, applying this to Qwen-Image-2512 yields a massive improvement, elevating its overall score from 42.83 to 58.39, with substantial gains across all logical dimensions. Taking this a step further, we explore the \textit{AutoFigure pipeline}~\cite{zhu2026autofigure}, an extreme form of reasoning scaling where Gemini 2.5 Flash~\cite{gemini2.5pro} explicitly writes SVG code to represent the framework. Evaluating this pure SVG output directly corroborates our earlier \textit{Findings 2} regarding the logic-aesthetic trade-off: the pure code-driven SVG achieves exceptional logical accuracy (\textit{e.g.}, 91.98 in Semantics) but suffers from severe aesthetic deficiencies (43.12). However, when this SVG is used as a structural prior to prompt a strong closed-source generator like Nano Banana Pro, we observe a synergistic effect, pushing the state-of-the-art overall score to 78.03.

\noindent\textbf{Scaling the Generation Phase.}
For models with robust inherent reasoning but imperfect rendering, we explore \textit{Post Enhancement} strategies at the generation end. We evaluate two specific pipelines on Wan2.6:
\textbf{1)} Best-of-N (BoN): We sample $N=4$ candidate images and utilize a strong VLM judge (Qwen3-VL-235B) to select the optimal output. This search strategy effectively filters out hallucinations, boosting the overall score from 65.84 to 67.23.
\textbf{2)} Post-Editing: Following~\cite{showtable}, we employ Nano Banana Pro as a dedicated editing model to explicitly correct localized visual artifacts and logical misalignments in the initial draft. This targeted refinement proves even more effective, elevating Wan2.6's performance to 69.31.
Ultimately, these explorations demonstrate that while end-to-end generation remains bottlenecked, strategically allocating compute at test-time, either to pre-plan the logical topology or post-refine the visual rendering, provides a highly effective roadmap for achieving production-ready academic illustrations.

\begin{findingbox}
    \textit{Findings 3:} Navigating the dual challenges of long, complex text reasoning and high-density rendering is crucial. By independently applying test time scaling to both stages, we can effectively shatter the capability ceilings of current models.
\end{findingbox}

\begin{figure*}[!t]
    \centering
    \includegraphics[width=1.0\linewidth]{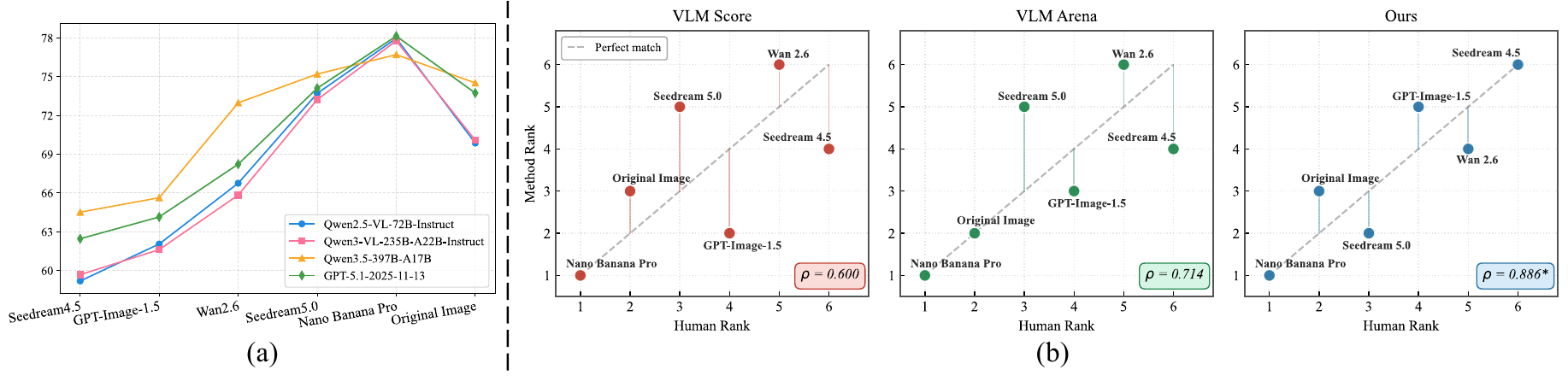}
    \vspace{-10pt}
    \caption{(a) shows the sensitivity analysis on different VLM solvers. (b) shows ranking correlation with human evaluation. $\rho$ denotes Spearman's rank correlation coefficient.}
    \label{fig:robust}
\end{figure*}

\subsection{Robustness Analysis of AIBench}
\noindent \textbf{Sensitivity of VLM Solvers.}
To investigate the robustness of our VQA-based evaluation framework, we analyze the sensitivity of the final scores to the choice of the VLM acting as the QA solver. As illustrated in Figure~\ref{fig:robust}(a), we evaluate the generated diagrams alongside the original images using four different SOTA VLMs (Qwen2.5-VL-72B~\cite{qwen2.5vl}, Qwen3-VL-235B~\cite{bai2025qwen3}, Qwen3.5-397B~\cite{qwen3}, and GPT-5.1~\cite{gpt5}). The absolute mean scores exhibit slight fluctuations across different solvers, which is an expected phenomenon stemming from the intrinsic ability in visual reasoning and instruction following. However, the relative performance ranking of the evaluated T2I models remains remarkably consistent. This demonstrates that AIBench provides a reliable assessment, ensuring that the benchmark's conclusions are robust and not biased by the VLM judge.


\noindent \textbf{Correlation with Human Evaluation.} To validate the effectiveness of our proposed metric, we conduct a system-level correlation analysis comparing our method, VLM-based judges, and human expert evaluation. For the VLM baselines, we employ Gemini 3 Flash and design two evaluation formats: 1) \textbf{VLM Score}: direct numerical scoring; 2) \textbf{VLM Arena}: a pairwise comparison approach (Original image \textit{vs} five generated). To establish a human rank, domain experts performed blind pairwise comparisons of generated images, with aggregated win rates forming the final ranking. 
As Figure~\ref{fig:robust}(b) illustrates, our method's rank correlation with human experts ($\rho = 0.886$) significantly exceeds the VLM Score ($\rho = 0.600$) and VLM Arena ($\rho = 0.714$) baselines, proving its better alignment with human intuition.

\section{Conclusion}
In this paper, we introduce AIBench, the first fine-grained, VQA-based benchmark specifically designed to evaluate the generation of complex academic illustrations. By moving beyond the unreliable paradigm of holistic VLM-as-Judge assessments, we establish a rigorous evaluation framework that explicitly decouples objective logical consistency from subjective aesthetic quality. Our extensive evaluations of state-of-the-art models reveal a profound capability gap when models are tasked with long, complex logical reasoning and high-density generation. Furthermore, we identify an inherent struggle of simultaneously optimizing rigorous structural fidelity and visual aesthetics. Finally, we demonstrate that test time scaling applied to either the reasoning or the generation phase can effectively shatter current capability ceilings. We hope AIBench serves as a critical stepping stone and roadmap for the community, driving future research toward multimodal generation frameworks that can master both deep logical planning and high-fidelity aesthetic rendering.
\clearpage
\bibliographystyle{plain}
\bibliography{main}

\begin{thebibliography}{10}

\bibitem{aesthetic_score}
Improved aesthetic predictor.
\newblock \url{https://github.com/christophschuhmann/improved-aesthetic-predictor}, 2022.

\bibitem{bai2025qwen3}
Shuai Bai, Yuxuan Cai, Ruizhe Chen, Keqin Chen, Xionghui Chen, Zesen Cheng, Lianghao Deng, Wei Ding, Chang Gao, Chunjiang Ge, et~al.
\newblock Qwen3-vl technical report.
\newblock {\em arXiv preprint arXiv:2511.21631}, 2025.

\bibitem{qwen2.5vl}
Shuai Bai, Keqin Chen, Xuejing Liu, Jialin Wang, Wenbin Ge, Sibo Song, Kai Dang, Peng Wang, Shijie Wang, Jun Tang, et~al.
\newblock Qwen2. 5-vl technical report.
\newblock {\em arXiv preprint arXiv:2502.13923}, 2025.

\bibitem{flux}
BlackForest.
\newblock Flux.
\newblock \url{https://github.com/black-forest-labs/flux}, 2024.

\bibitem{cao2025unipercept}
Shuo Cao, Jiayang Li, Xiaohui Li, Yuandong Pu, Kaiwen Zhu, Yuanting Gao, Siqi Luo, Yi~Xin, Qi~Qin, Yu~Zhou, et~al.
\newblock Unipercept: Towards unified perceptual-level image understanding across aesthetics, quality, structure, and texture.
\newblock {\em arXiv preprint arXiv:2512.21675}, 2025.

\bibitem{chen2024textdiffuser}
Jingye Chen, Yupan Huang, Tengchao Lv, Lei Cui, Qifeng Chen, and Furu Wei.
\newblock Textdiffuser-2: Unleashing the power of language models for text rendering.
\newblock In {\em European Conference on Computer Vision}, pages 386--402. Springer, 2024.

\bibitem{blip3o}
Jiuhai Chen, Zhiyang Xu, Xichen Pan, Yushi Hu, Can Qin, Tom Goldstein, Lifu Huang, Tianyi Zhou, Saining Xie, Silvio Savarese, et~al.
\newblock Blip3-o: A family of fully open unified multimodal models-architecture, training and dataset.
\newblock {\em arXiv preprint arXiv:2505.09568}, 2025.

\bibitem{chen2025blip3onext}
Jiuhai Chen, Le~Xue, Zhiyang Xu, Xichen Pan, Shusheng Yang, Can Qin, An~Yan, Honglu Zhou, Zeyuan Chen, Lifu Huang, et~al.
\newblock Blip3o-next: Next frontier of native image generation.
\newblock {\em arXiv preprint arXiv:2510.15857}, 2025.

\bibitem{januspro}
Xiaokang Chen, Zhiyu Wu, Xingchao Liu, Zizheng Pan, Wen Liu, Zhenda Xie, Xingkai Yu, and Chong Ruan.
\newblock Janus-pro: Unified multimodal understanding and generation with data and model scaling.
\newblock {\em arXiv preprint arXiv:2501.17811}, 2025.

\bibitem{gemini2.5pro}
Gheorghe Comanici, Eric Bieber, Mike Schaekermann, Ice Pasupat, Noveen Sachdeva, Inderjit Dhillon, Marcel Blistein, Ori Ram, Dan Zhang, Evan Rosen, et~al.
\newblock Gemini 2.5: Pushing the frontier with advanced reasoning, multimodality, long context, and next generation agentic capabilities.
\newblock {\em arXiv preprint arXiv:2507.06261}, 2025.

\bibitem{cui2025emu3}
Yufeng Cui, Honghao Chen, Haoge Deng, Xu~Huang, Xinghang Li, Jirong Liu, Yang Liu, Zhuoyan Luo, Jinsheng Wang, Wenxuan Wang, et~al.
\newblock Emu3.5: Native multimodal models are world learners.
\newblock {\em arXiv preprint arXiv:2510.26583}, 2025.

\bibitem{bagel}
Chaorui Deng, Deyao Zhu, Kunchang Li, Chenhui Gou, Feng Li, Zeyu Wang, Shu Zhong, Weihao Yu, Xiaonan Nie, Ziang Song, et~al.
\newblock Emerging properties in unified multimodal pretraining.
\newblock {\em arXiv preprint arXiv:2505.14683}, 2025.

\bibitem{sd3}
Patrick Esser, Sumith Kulal, Andreas Blattmann, Rahim Entezari, Jonas M{\"u}ller, Harry Saini, Yam Levi, Dominik Lorenz, Axel Sauer, Frederic Boesel, et~al.
\newblock Scaling rectified flow transformers for high-resolution image synthesis.
\newblock In {\em Forty-first international conference on machine learning}, 2024.

\bibitem{postermaker}
Yifan Gao, Zihang Lin, Chuanbin Liu, Min Zhou, Tiezheng Ge, Bo~Zheng, and Hongtao Xie.
\newblock {PosterMaker}: Towards high-quality product poster generation with accurate text rendering.
\newblock In {\em CVPR}, pages 8083--8093, 2025.

\bibitem{ghosh2023geneval}
Dhruba Ghosh, Hannaneh Hajishirzi, and Ludwig Schmidt.
\newblock {GenEval}: An object-focused framework for evaluating text-to-image alignment.
\newblock In {\em {Advances in Neural Inf. Process. Syst.}}, pages 52132--52152, 2023.

\bibitem{gemini3}
Google.
\newblock gemini3.
\newblock \url{https://blog.google/products-and-platforms/products/gemini/gemini-3/}, 2025.

\bibitem{nano_banana}
Google.
\newblock Nano banana pro.
\newblock \url{https://blog.google/innovation-and-ai/products/nano-banana-pro/}, 2025.

\bibitem{dreamposter}
Xiwei Hu, Haokun Chen, Zhongqi Qi, Hui Zhang, Dexiang Hong, Jie Shao, and Xinglong Wu.
\newblock {DreamPoster}: A unified framework for image-conditioned generative poster design.
\newblock {\em arXiv preprint arXiv:2507.04218}, 2025.

\bibitem{PPSGen}
Yue Hu and Xiaojun Wan.
\newblock {PPSGen}: Learning-based presentation slides generation for academic papers.
\newblock In {\em IJCAI}, pages 1085--1097, 2014.

\bibitem{huang2025t2i}
Kaiyi Huang, Chengqi Duan, Kaiyue Sun, Enze Xie, Zhenguo Li, and Xihui Liu.
\newblock T2i-compbench++: An enhanced and comprehensive benchmark for compositional text-to-image generation.
\newblock {\em {{IEEE} Trans. Pattern Anal. Mach. Intell.}}, 47(5):3563--3579, 2025.

\bibitem{t2i_r1}
Dongzhi Jiang, Ziyu Guo, Renrui Zhang, Zhuofan Zong, Hao Li, Le~Zhuo, Shilin Yan, Pheng-Ann Heng, and Hongsheng Li.
\newblock {T2I-R1}: Reinforcing image generation with collaborative semantic-level and token-level cot.
\newblock {\em arXiv preprint arXiv:2505.00703}, 2025.

\bibitem{jiang2025enhancing}
Kaixun Jiang, Zhaoyu Chen, Haijing Guo, Jinglun Li, Jiyuan Fu, Pinxue Guo, Hao Tang, Bo~Li, and Wenqiang Zhang.
\newblock Enhancing diffusion-based unrestricted adversarial attacks via adversary preferences alignment.
\newblock {\em arXiv preprint arXiv:2506.01511}, 2025.

\bibitem{jiang2026genagent}
Kaixun Jiang, Yuzheng Wang, Junjie Zhou, Pandeng Li, Zhihang Liu, Chen-Wei Xie, Zhaoyu Chen, Yun Zheng, and Wenqiang Zhang.
\newblock Genagent: Scaling text-to-image generation via agentic multimodal reasoning.
\newblock {\em arXiv preprint arXiv:2601.18543}, 2026.

\bibitem{kirstain2023pick}
Yuval Kirstain, Adam Polyak, Uriel Singer, Shahbuland Matiana, Joe Penna, and Omer Levy.
\newblock {Pick-a-Pic}: An open dataset of user preferences for text-to-image generation.
\newblock In {\em {Advances in Neural Inf. Process. Syst.}}, pages 36652--36663, 2023.

\bibitem{uniworld}
Bin Lin, Zongjian Li, Xinhua Cheng, Yuwei Niu, Yang Ye, Xianyi He, Shenghai Yuan, Wangbo Yu, Shaodong Wang, Yunyang Ge, et~al.
\newblock {UniWorld}: High-resolution semantic encoders for unified visual understanding and generation.
\newblock {\em arXiv preprint arXiv:2506.03147}, 2025.

\bibitem{autoposter}
Jinpeng Lin, Min Zhou, Ye~Ma, Yifan Gao, Chenxi Fei, Yangjian Chen, Zhang Yu, and Tiezheng Ge.
\newblock {AutoPoster}: A highly automatic and content-aware design system for advertising poster generation.
\newblock In {\em ACM MM}, pages 1250--1260, 2023.

\bibitem{aesR1}
Boyang Liu, Yifan Hu, Senjie Jin, Shihan Dou, Gonglei Shi, Jie Shao, Tao Gui, and Xuanjing Huang.
\newblock Unlocking the essence of beauty: Advanced aesthetic reasoning with relative-absolute policy optimization.
\newblock {\em arXiv preprint arXiv:2509.21871}, 2025.

\bibitem{liu2024glyph}
Zeyu Liu, Weicong Liang, Zhanhao Liang, Chong Luo, Ji~Li, Gao Huang, and Yuhui Yuan.
\newblock {Glyph-ByT5}: A customized text encoder for accurate visual text rendering.
\newblock In {\em European Conference on Computer Vision}, pages 361--377. Springer, 2024.

\bibitem{showtable}
Zhihang Liu, Xiaoyi Bao, Pandeng Li, Junjie Zhou, Zhaohe Liao, Yefei He, Kaixun Jiang, Chen-Wei Xie, Yun Zheng, and Hongtao Xie.
\newblock {ShowTable}: Unlocking creative table visualization with collaborative reflection and refinement.
\newblock {\em arXiv preprint arXiv:2512.13303}, 2025.

\bibitem{liu2025hybrid}
Zhihang Liu, Chen-Wei Xie, Pandeng Li, Liming Zhao, Longxiang Tang, Yun Zheng, Chuanbin Liu, and Hongtao Xie.
\newblock Hybrid-level instruction injection for video token compression in multi-modal large language models.
\newblock In {\em Proceedings of the Computer Vision and Pattern Recognition Conference}, pages 8568--8578, 2025.

\bibitem{liu2025capability}
Zhihang Liu, Chen-Wei Xie, Bin Wen, Feiwu Yu, Jixuan Chen, Pandeng Li, Boqiang Zhang, Nianzu Yang, Yinglu Li, Zuan Gao, Yun Zheng, and Hongtao Xie.
\newblock Capability: A comprehensive visual caption benchmark for evaluating both correctness and thoroughness, 2025.

\bibitem{niu2025wise}
Yuwei Niu, Munan Ning, Mengren Zheng, Weiyang Jin, Bin Lin, Peng Jin, Jiaqi Liao, Chaoran Feng, Kunpeng Ning, Bin Zhu, et~al.
\newblock Wise: A world knowledge-informed semantic evaluation for text-to-image generation.
\newblock {\em arXiv preprint arXiv:2503.07265}, 2025.

\bibitem{dalle3}
OpenAI.
\newblock Dall·e 3.
\newblock \url{https://openai.com/index/dall-e-3/}, 2023.

\bibitem{gpt5}
OpenAI.
\newblock Gpt-5.
\newblock \url{https://openai.com/index/introducing-gpt-5/}, 2025.

\bibitem{gpt-image-1.5}
OpenAI.
\newblock Gpt-image-1.5.
\newblock \url{https://openai.com/index/new-chatgpt-images-is-here/}, 2025.

\bibitem{pang2025paper2poster}
Wei Pang, Kevin~Qinghong Lin, Xiangru Jian, Xi~He, and Philip Torr.
\newblock {Paper2poster}: Towards multimodal poster automation from scientific papers.
\newblock {\em arXiv preprint arXiv:2505.21497}, 2025.

\bibitem{qiang2016learning}
Yuting Qiang, Yanwei Fu, Yanwen Guo, Zhi-Hua Zhou, and Leonid Sigal.
\newblock Learning to generate posters of scientific papers.
\newblock In {\em AAAI}, pages 929--937, 2016.

\bibitem{sd}
Robin Rombach, Andreas Blattmann, Dominik Lorenz, Patrick Esser, and Bj{\"o}rn Ommer.
\newblock High-resolution image synthesis with latent diffusion models.
\newblock In {\em Proceedings of the IEEE/CVF conference on computer vision and pattern recognition}, pages 10684--10695, 2022.

\bibitem{seedream-5-0-lite}
ByteDance Seed.
\newblock Seedream 5.0 lite.
\newblock \url{https://seed.bytedance.com/en/blog/deeper-thinking-more-accurate-generation-introducing-seedream-5-0-lite}, 2025.

\bibitem{seedance15pro}
Team Seedance, Heyi Chen, Siyan Chen, Xin Chen, Yanfei Chen, Ying Chen, Zhuo Chen, Feng Cheng, Tianheng Cheng, Xinqi Cheng, Xuyan Chi, Jian Cong, Jing Cui, Qinpeng Cui, Qide Dong, Junliang Fan, Jing Fang, Zetao Fang, Chengjian Feng, Han Feng, Mingyuan Gao, Yu~Gao, Dong Guo, Qiushan Guo, Boyang Hao, Qingkai Hao, Bibo He, Qian He, Tuyen Hoang, Ruoqing Hu, Xi~Hu, Weilin Huang, Zhaoyang Huang, Zhongyi Huang, Donglei Ji, Siqi Jiang, Wei Jiang, Yunpu Jiang, Zhuo Jiang, Ashley Kim, Jianan Kong, Zhichao Lai, Shanshan Lao, Yichong Leng, Ai~Li, Feiya Li, Gen Li, Huixia Li, JiaShi Li, Liang Li, Ming Li, Shanshan Li, Tao Li, Xian Li, Xiaojie Li, Xiaoyang Li, Xingxing Li, Yameng Li, Yifu Li, Yiying Li, Chao Liang, Han Liang, Jianzhong Liang, Ying Liang, Zhiqiang Liang, Wang Liao, Yalin Liao, Heng Lin, Kengyu Lin, Shanchuan Lin, Xi~Lin, Zhijie Lin, Feng Ling, Fangfang Liu, Gaohong Liu, Jiawei Liu, Jie Liu, Jihao Liu, Shouda Liu, Shu Liu, Sichao Liu, Songwei Liu, Xin Liu, Xue Liu, Yibo Liu, Zikun Liu, Zuxi Liu, Junlin Lyu,
  Lecheng Lyu, Qian Lyu, Han Mu, Xiaonan Nie, Jingzhe Ning, Xitong Pan, Yanghua Peng, Lianke Qin, Xueqiong Qu, Yuxi Ren, Kai Shen, Guang Shi, Lei Shi, Yan Song, Yinglong Song, Fan Sun, Li~Sun, Renfei Sun, Yan Sun, Zeyu Sun, Wenjing Tang, Yaxue Tang, Zirui Tao, Feng Wang, Furui Wang, Jinran Wang, Junkai Wang, Ke~Wang, Kexin Wang, Qingyi Wang, Rui Wang, Sen Wang, Shuai Wang, Tingru Wang, Weichen Wang, Xin Wang, Yanhui Wang, Yue Wang, Yuping Wang, Yuxuan Wang, Ziyu Wang, Guoqiang Wei, Wanru Wei, Di~Wu, Guohong Wu, Hanjie Wu, Jian Wu, Jie Wu, Ruolan Wu, Xinglong Wu, Yonghui Wu, Ruiqi Xia, Liang Xiang, Fei Xiao, XueFeng Xiao, Pan Xie, Shuangyi Xie, Shuang Xu, Jinlan Xue, Shen Yan, Bangbang Yang, Ceyuan Yang, Jiaqi Yang, Runkai Yang, Tao Yang, Yang Yang, Yihang Yang, ZhiXian Yang, Ziyan Yang, Songting Yao, Yifan Yao, Zilyu Ye, Bowen Yu, Jian Yu, Chujie Yuan, Linxiao Yuan, Sichun Zeng, Weihong Zeng, Xuejiao Zeng, Yan Zeng, Chuntao Zhang, Heng Zhang, Jingjie Zhang, Kuo Zhang, Liang Zhang, Liying Zhang, Manlin Zhang,
  Ting Zhang, Weida Zhang, Xiaohe Zhang, Xinyan Zhang, Yan Zhang, Yuan Zhang, Zixiang Zhang, Fengxuan Zhao, Huating Zhao, Yang Zhao, Hao Zheng, Jianbin Zheng, Xiaozheng Zheng, Yangyang Zheng, Yijie Zheng, Jiexin Zhou, Jiahui Zhu, Kuan Zhu, Shenhan Zhu, Wenjia Zhu, Benhui Zou, and Feilong Zuo.
\newblock Seedance 1.5 pro: A native audio-visual joint generation foundation model, 2025.

\bibitem{seedream2025seedream}
Team Seedream, Yunpeng Chen, Yu~Gao, Lixue Gong, Meng Guo, Qiushan Guo, Zhiyao Guo, Xiaoxia Hou, Weilin Huang, Yixuan Huang, et~al.
\newblock Seedream 4.0: Toward next-generation multimodal image generation.
\newblock {\em arXiv preprint arXiv:2509.20427}, 2025.

\bibitem{kline-omni}
Kling Team, Jialu Chen, Yuanzheng Ci, Xiangyu Du, Zipeng Feng, Kun Gai, Sainan Guo, Feng Han, Jingbin He, Kang He, Xiao Hu, Xiaohua Hu, Boyuan Jiang, Fangyuan Kong, Hang Li, Jie Li, Qingyu Li, Shen Li, Xiaohan Li, Yan Li, Jiajun Liang, Borui Liao, Yiqiao Liao, Weihong Lin, Quande Liu, Xiaokun Liu, Yilun Liu, Yuliang Liu, Shun Lu, Hangyu Mao, Yunyao Mao, Haodong Ouyang, Wenyu Qin, Wanqi Shi, Xiaoyu Shi, Lianghao Su, Haozhi Sun, Peiqin Sun, Pengfei Wan, Chao Wang, Chenyu Wang, Meng Wang, Qiulin Wang, Runqi Wang, Xintao Wang, Xuebo Wang, Zekun Wang, Min Wei, Tiancheng Wen, Guohao Wu, Xiaoshi Wu, Zhenhua Wu, Da~Xie, Yingtong Xiong, Yulong Xu, Sile Yang, Zikang Yang, Weicai Ye, Ziyang Yuan, Shenglong Zhang, Shuaiyu Zhang, Yuanxing Zhang, Yufan Zhang, Wenzheng Zhao, Ruiliang Zhou, Yan Zhou, Guosheng Zhu, and Yongjie Zhu.
\newblock Kling-omni technical report, 2025.

\bibitem{wan26}
Wan Team.
\newblock Wan2.6.
\newblock \url{https://wan.video/introduction/wan2.6}, 2025.

\bibitem{team2025zimage}
Z-Image Team.
\newblock Z-image: An efficient image generation foundation model with single-stream diffusion transformer.
\newblock {\em arXiv preprint arXiv:2511.22699}, 2025.

\bibitem{tuo2024anytext2}
Yuxiang Tuo, Yifeng Geng, and Liefeng Bo.
\newblock Anytext2: Visual text generation and editing with customizable attributes.
\newblock {\em arXiv preprint arXiv:2411.15245}, 2024.

\bibitem{emu3}
Xinlong Wang, Xiaosong Zhang, Zhengxiong Luo, Quan Sun, Yufeng Cui, Jinsheng Wang, Fan Zhang, Yueze Wang, Zhen Li, Qiying Yu, et~al.
\newblock Emu3: Next-token prediction is all you need.
\newblock {\em arXiv preprint arXiv:2409.18869}, 2024.

\bibitem{wei2025dreamrelation}
Yujie Wei, Shiwei Zhang, Hangjie Yuan, Biao Gong, Longxiang Tang, Xiang Wang, Haonan Qiu, Hengjia Li, Shuai Tan, Yingya Zhang, et~al.
\newblock Dreamrelation: Relation-centric video customization.
\newblock In {\em Proceedings of the IEEE/CVF International Conference on Computer Vision}, pages 12381--12393, 2025.

\bibitem{wei2025routing}
Yujie Wei, Shiwei Zhang, Hangjie Yuan, Yujin Han, Zhekai Chen, Jiayu Wang, Difan Zou, Xihui Liu, Yingya Zhang, Yu~Liu, et~al.
\newblock Routing matters in moe: Scaling diffusion transformers with explicit routing guidance.
\newblock {\em arXiv preprint arXiv:2510.24711}, 2025.

\bibitem{qwenimage}
Chenfei Wu, Jiahao Li, Jingren Zhou, Junyang Lin, Kaiyuan Gao, Kun Yan, Sheng-ming Yin, Shuai Bai, Xiao Xu, Yilei Chen, et~al.
\newblock Qwen-image technical report.
\newblock {\em arXiv preprint arXiv:2508.02324}, 2025.

\bibitem{omnigen2}
Chenyuan Wu, Pengfei Zheng, Ruiran Yan, Shitao Xiao, Xin Luo, Yueze Wang, Wanli Li, Xiyan Jiang, Yexin Liu, Junjie Zhou, et~al.
\newblock Omnigen2: Exploration to advanced multimodal generation.
\newblock {\em arXiv preprint arXiv:2506.18871}, 2025.

\bibitem{showo}
Jinheng Xie, Weijia Mao, Zechen Bai, David~Junhao Zhang, Weihao Wang, Kevin~Qinghong Lin, Yuchao Gu, Zhijie Chen, Zhenheng Yang, and Mike~Zheng Shou.
\newblock Show-o: One single transformer to unify multimodal understanding and generation.
\newblock {\em arXiv preprint arXiv:2408.12528}, 2024.

\bibitem{xu2022posterbot}
Sheng Xu and Xiaojun Wan.
\newblock Posterbot: A system for generating posters of scientific papers with neural models.
\newblock In {\em AAAI}, pages 13233--13235, 2022.

\bibitem{qwen3}
An~Yang, Anfeng Li, Baosong Yang, Beichen Zhang, Binyuan Hui, Bo~Zheng, Bowen Yu, Chang Gao, Chengen Huang, Chenxu Lv, et~al.
\newblock Qwen3 technical report.
\newblock {\em arXiv preprint arXiv:2505.09388}, 2025.

\bibitem{PPTAGENT}
Hao Zheng, Xinyan Guan, Hao Kong, Wenkai Zhang, Jia Zheng, Weixiang Zhou, Hongyu Lin, Yaojie Lu, Xianpei Han, and Le~Sun.
\newblock {PPTAGENT}: Generating and evaluating presentations beyond text-to-slides.
\newblock In {\em EMNLP}, pages 14413--14429, 2025.

\bibitem{zhu2026paperbanana}
Dawei Zhu, Rui Meng, Yale Song, Xiyu Wei, Sujian Li, Tomas Pfister, and Jinsung Yoon.
\newblock {PaperBanana}: Automating academic illustration for ai scientists.
\newblock {\em arXiv preprint arXiv:2601.23265}, 2026.

\bibitem{zhu2025internvl3}
Jinguo Zhu, Weiyun Wang, Zhe Chen, Zhaoyang Liu, Shenglong Ye, Lixin Gu, Hao Tian, Yuchen Duan, Weijie Su, Jie Shao, et~al.
\newblock Internvl3: Exploring advanced training and test-time recipes for open-source multimodal models.
\newblock {\em arXiv preprint arXiv:2504.10479}, 2025.

\bibitem{zhu2026autofigure}
Minjun Zhu, Zhen Lin, Yixuan Weng, Panzhong Lu, Qiujie Xie, Yifan Wei, Sifan Liu, Qiyao Sun, and Yue Zhang.
\newblock {AutoFigure}: Generating and refining publication-ready scientific illustrations.
\newblock {\em arXiv preprint arXiv:2602.03828}, 2026.

\end{thebibliography}


\appendix
\clearpage

\providecommand{\authcount}[1]{}

\newcommand\DoToC{%
    \hypersetup{linkcolor=eccvblue}
    \setcounter{tocdepth}{2}
    \startcontents
    \printcontents{}{1}{\hrulefill\vskip0pt}
    \vskip0pt \noindent\hrulefill
    }

\setcounter{page}{1}
\setcounter{table}{0}
\setcounter{figure}{0}
\setcounter{equation}{0}
\setcounter{footnote}{0}
\renewcommand{\thetable}{A\arabic{table}}
\renewcommand{\thefigure}{A\arabic{figure}}
\renewcommand{\theequation}{A\arabic{equation}}

\begin{center}
    \Large
    \textbf{Appendix}
    \vspace{1.0em}
\end{center}

\noindent\textbf{Overview}
\noindent\DoToC

\section{More Details of AIBench}

\subsection{QA Construction}

\begin{figure*}[!t]
    \centering
    \includegraphics[width=1.0\linewidth]{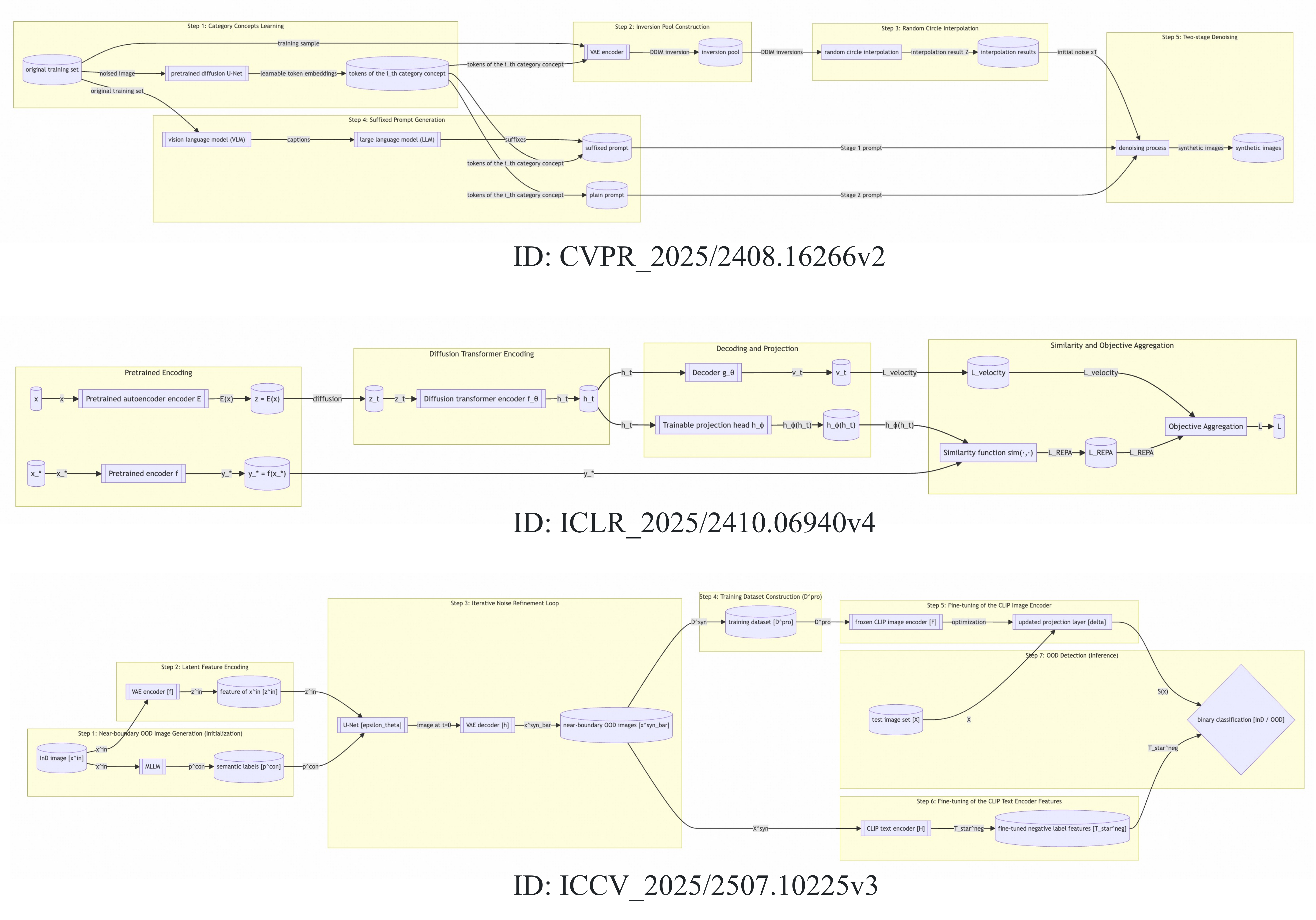}
    \vspace{-10pt}
    \caption{Examples of rendered text-to-logic directed graph derived from method-text. Purple boxes of two shapes represent nodes: rectangular purple boxes denote processing modules (\textit{e.g.}, VAE encoder, MLLM), while elliptical purple boxes represent data entities (\textit{e.g.}, InD images, latent features). Black arrows denote edges, indicating the direction of data flow between nodes. Yellow boxes indicate phases, corresponding to distinct procedural steps (\textit{e.g.}, Image Generation, Feature Encoding).}
    \label{fig:graph_json}
    \vspace{-1.3em}
\end{figure*}

\begin{figure*}[!t]
    \centering
    \includegraphics[width=1.0\linewidth]{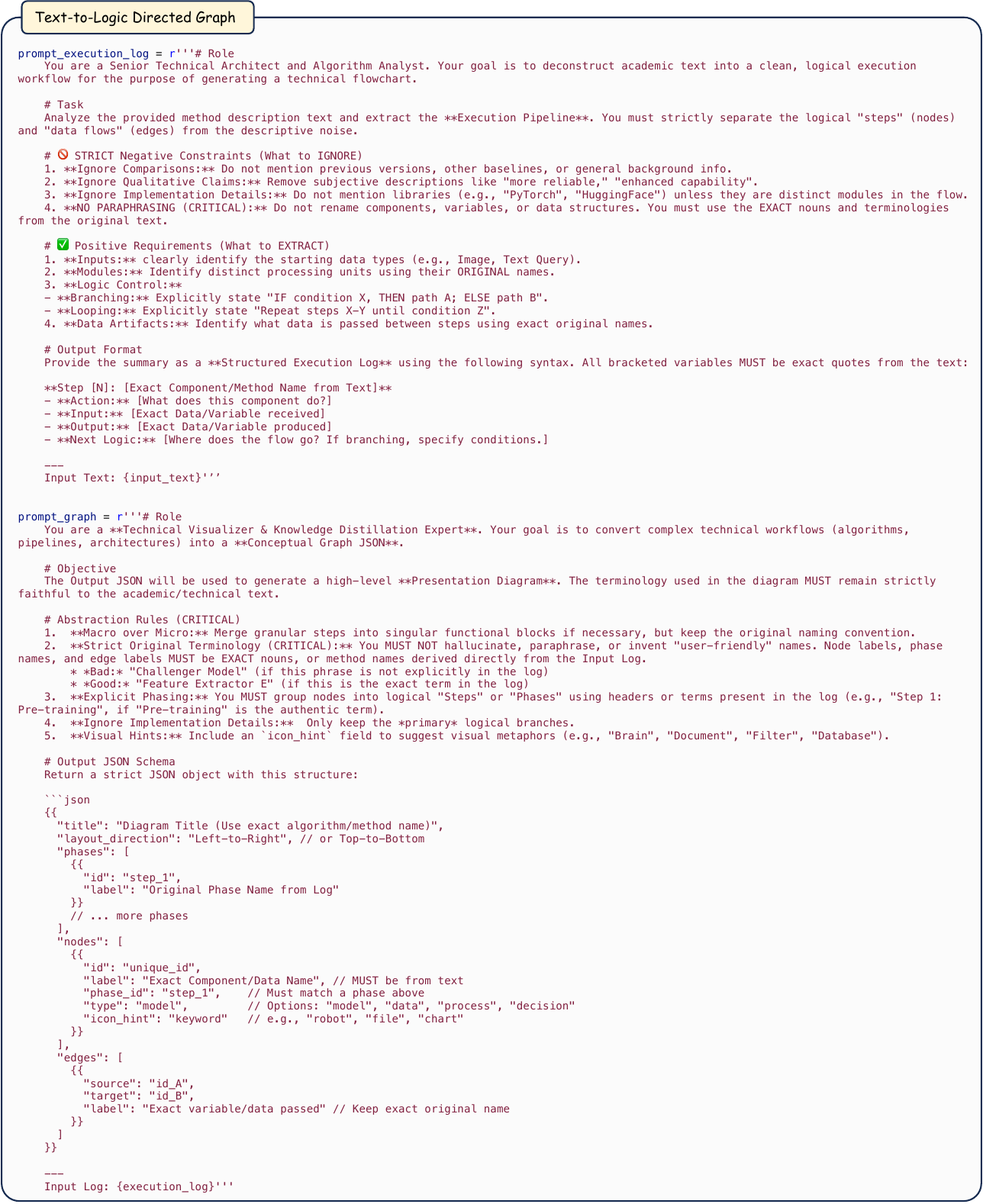}
    \vspace{-10pt}
    \caption{Prompt of Text-to-Logic Directed Graph during the QA construction.}
    \label{fig:Text-to-Logic Directed Graph}
\end{figure*}

\noindent \textbf{Text-to-Logic Directed Graph.} The construction of the Text-to-Logic Directed Graph proceeds in two stages: we first extract the core logic from the raw method text, and then generate the Logic Directed Graph based on the extracted execution logs. This process is implemented using the Gemini 3 Flash model, with the corresponding prompts illustrated in Figure~\ref{fig:Text-to-Logic Directed Graph}. We also render several cases of directed graphs in Figure~\ref{fig:graph_json}.

\begin{figure*}[!t]
    \centering
    \includegraphics[width=1.0\linewidth]{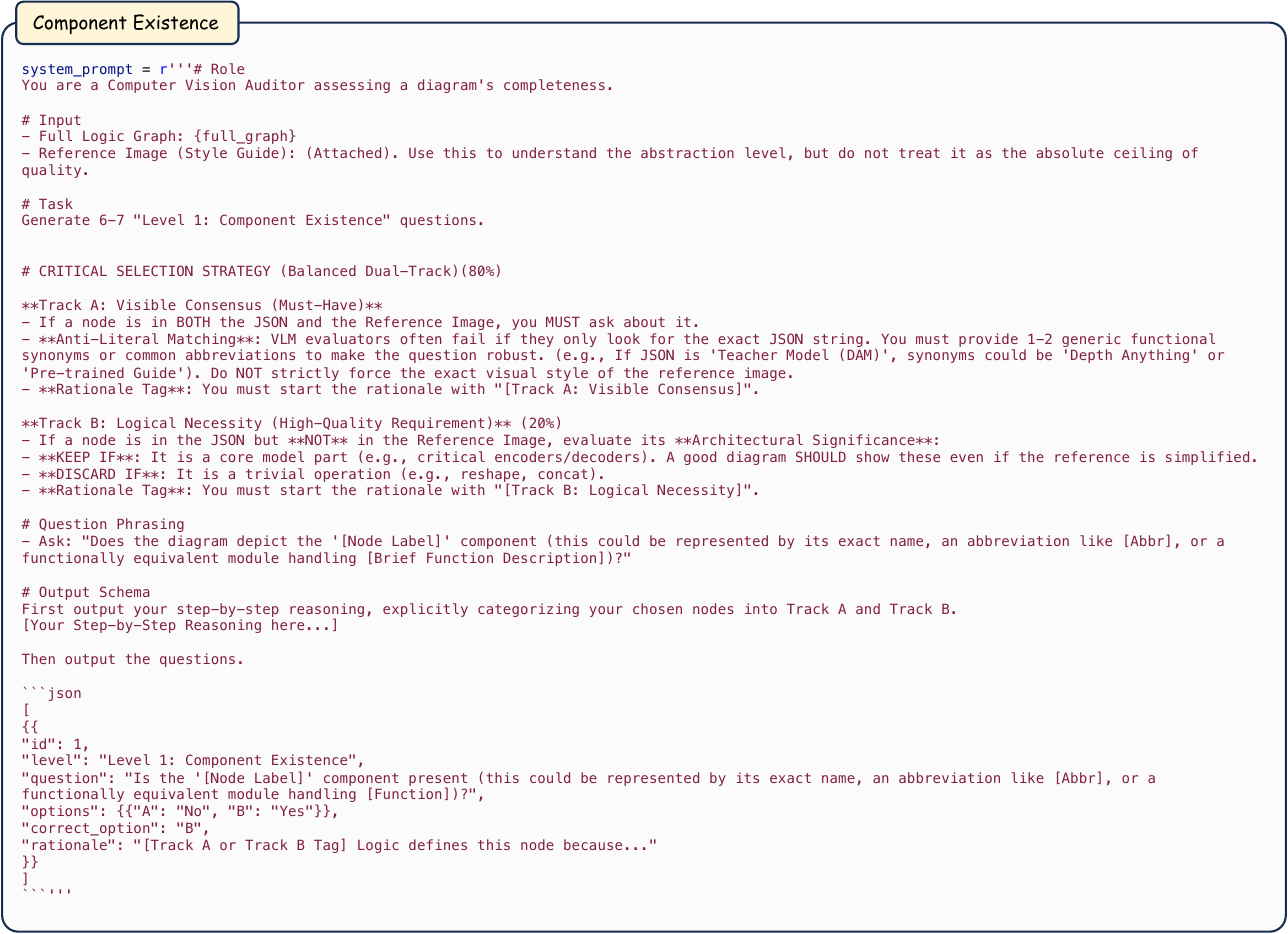}
    \vspace{-10pt}
    \caption{Prompt of Level 1: Component Existence during the QA construction.}
    \label{fig:Component Existence}
\end{figure*}

\begin{figure*}[!t]
    \centering
    \includegraphics[width=0.95\linewidth]{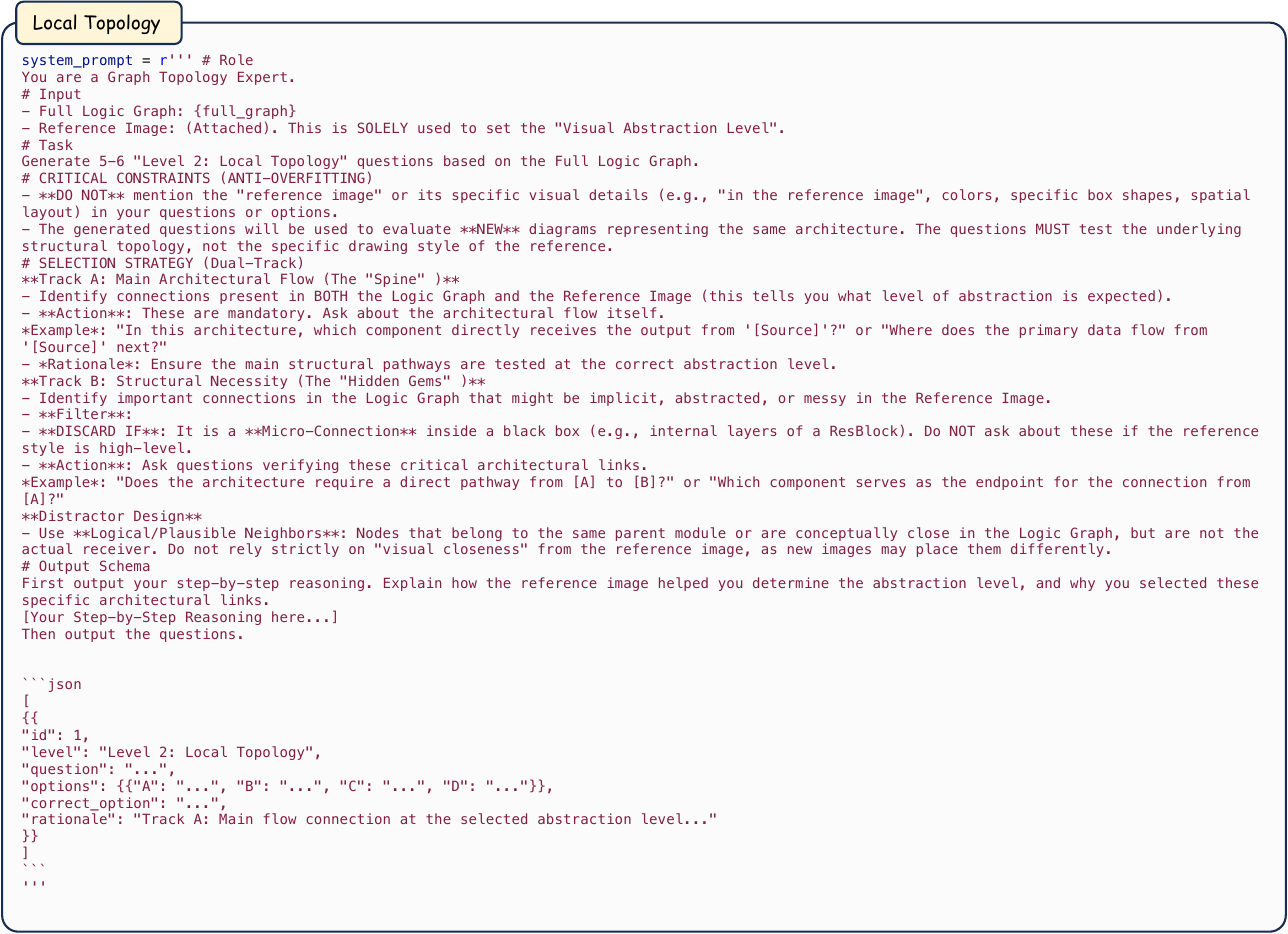}
    \caption{Prompt of Level 2: Local Topology during the QA construction.}
    \label{fig:Local Topology}
\end{figure*}

\begin{figure*}[!t]
    \centering
    \includegraphics[width=0.95\linewidth]{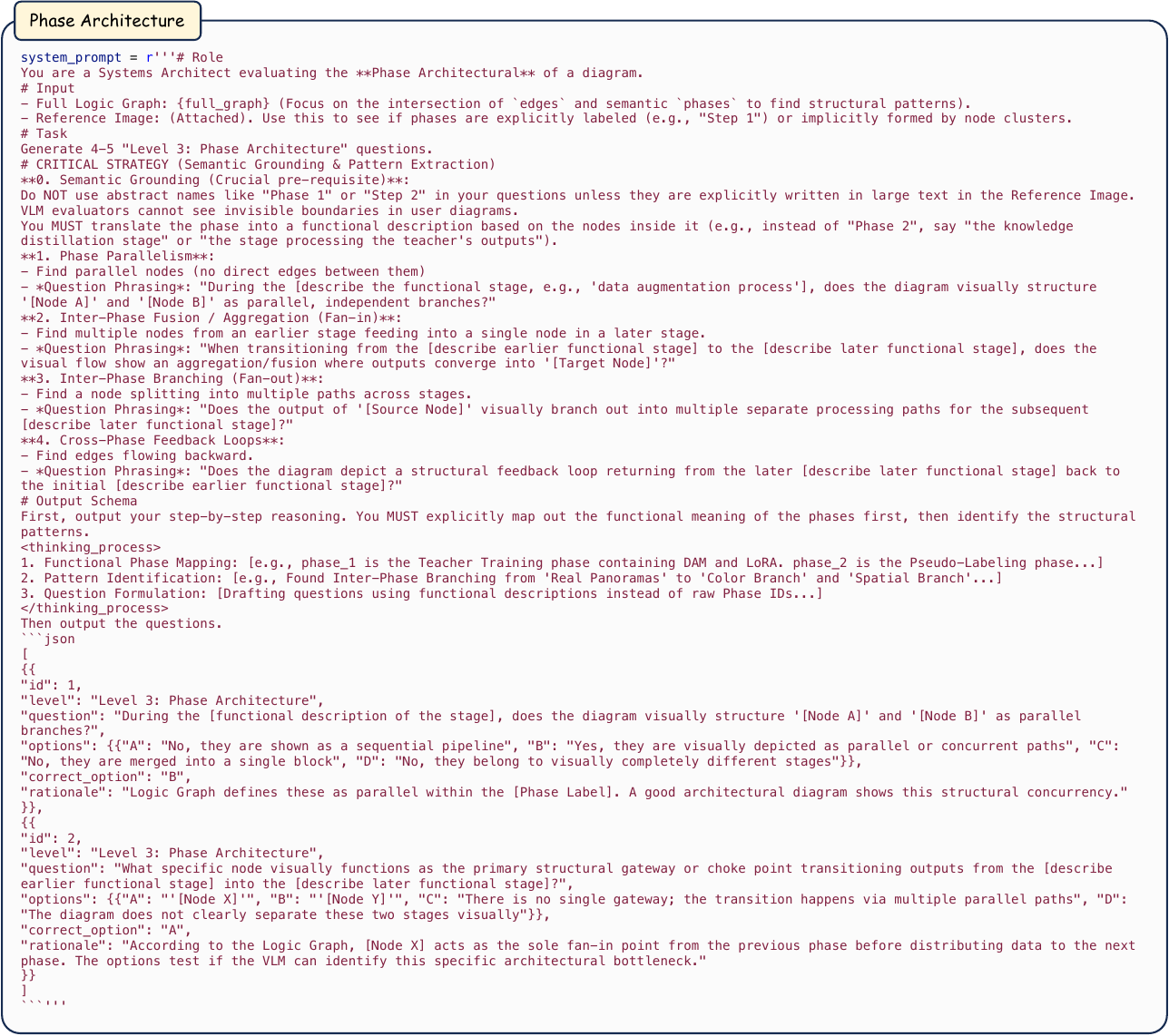}
    \caption{Prompt of Level 3: Phase Architecture during the QA construction.}
    \label{fig:Phase Architecture}
\end{figure*}

\begin{figure*}[!t]
    \centering
    \includegraphics[width=1.0\linewidth]{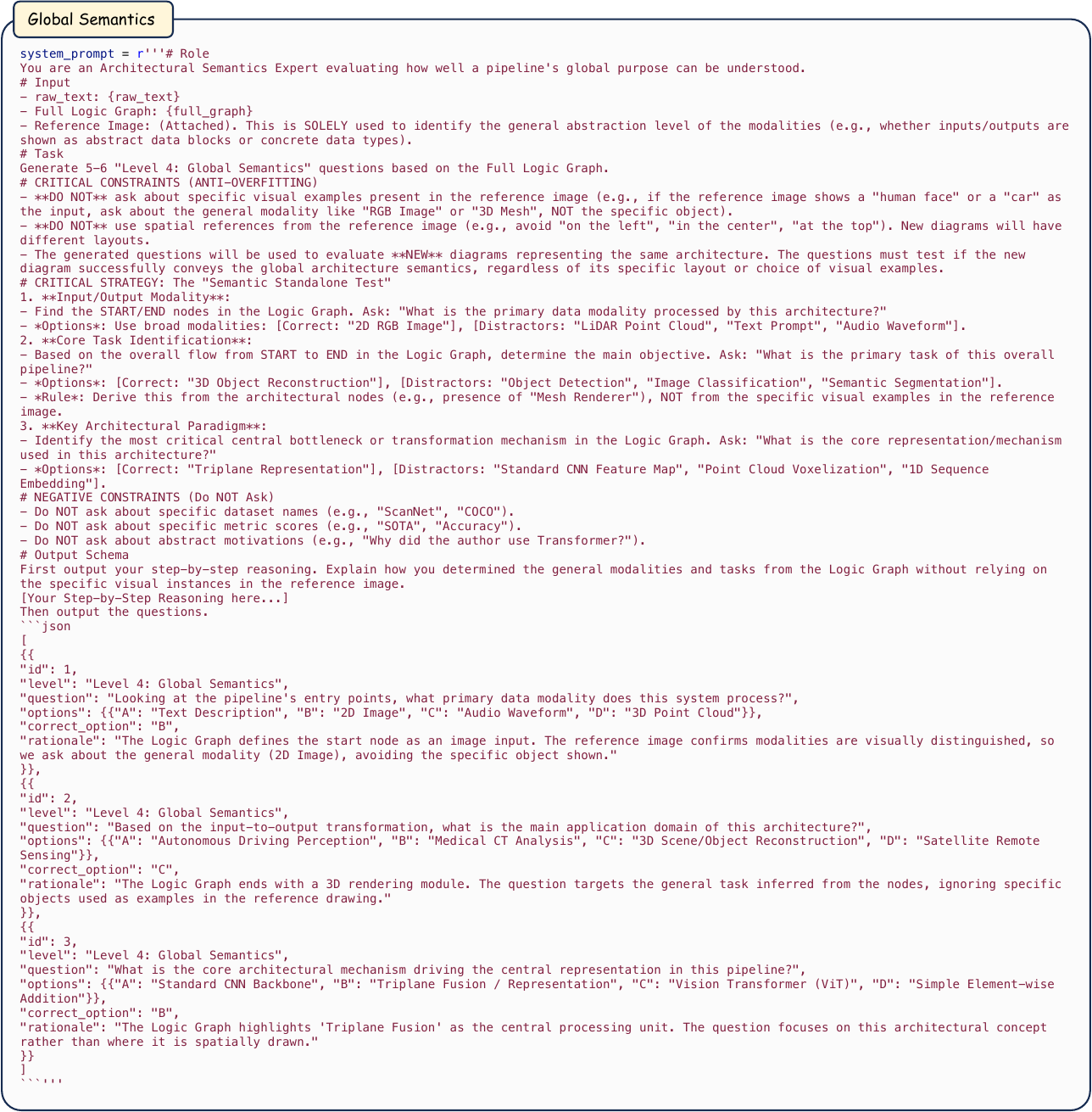}
    \vspace{-10pt}
    \caption{Prompt of Level 4: Global Semantics during the QA construction.}
    \label{fig:Global Semantics}
\end{figure*}

\noindent \textbf{Multi-Level QA Generation.} We utilize Gemini 3 Flash to generate QA pairs across four levels. The prompt designs for each level are illustrated in the figures provided: {Level 1: Component Existence} (Figure~\ref{fig:Component Existence}), {Level 2: Local Topology} (Figure~\ref{fig:Local Topology}), {Level 3: Phase Architecture} (Figure~\ref{fig:Phase Architecture}), {Level 4: Global Semantics} (Figure~\ref{fig:Global Semantics}).

\subsection{Human Expert Annotation Details}
To ensure the generation of high-quality and reliable ground truth data, a rigorous manual annotation process was conducted by human experts. All selected annotators are Ph.D. students possessing relevant domain expertise, specifically the first seven authors of this paper. In total, 300 articles were meticulously annotated. With each sample requiring an average of 20 minutes of thorough review, the entire annotation effort encompassed a substantial workload of approximately 100 hours of dedicated expert labor.

\section{More Experiments}
\subsection{More Experimental Settings and Details}

\begin{figure*}[t]
    \centering
    \includegraphics[width=1.0\linewidth]{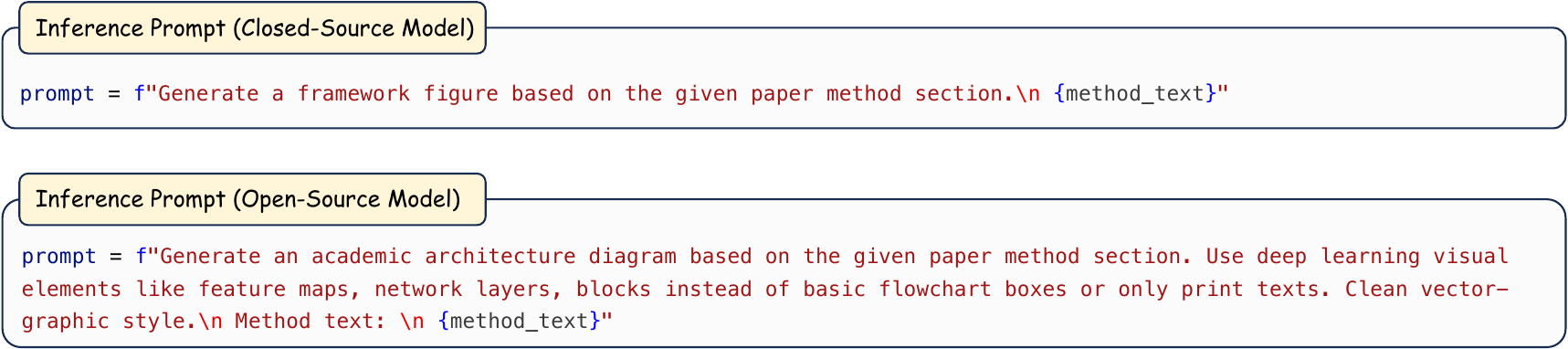}
    \vspace{-10pt}
    \caption{Different inference prompts for closed- and open-source models.}
    \label{fig:infer_prompt}
    \vspace{-1.3em}
\end{figure*}

\noindent\textbf{Model Inference and Prompt Design.} 
To ensure a fair and comprehensive evaluation, we separately design the inference prompts for closed- and open-source models tailored to the distinct capabilities of the evaluated models, as shown in Fig.~\ref{fig:infer_prompt}. Because closed-source models generally exhibit superior instruction-following and long-text comprehension capabilities, we utilize a relatively concise prompt for this group. Conversely, current open-source models often struggle to grasp the specific structural requirements of academic illustration generation. Without sufficient constraints, these models frequently exhibit degenerate behaviors, such as directly rendering the raw methodology text onto the image canvas instead of synthesizing a visual framework. To mitigate this, we employ a more detailed and explicitly constrained prompt for the open-source models. Importantly, to prevent trivial outputs across the board, the prompts for all models are carefully adjusted to include negative constraints. Specifically, the models were explicitly instructed not to merely print the raw text and not to generate overly simplistic, basic flowcharts.

\begin{figure*}[t]
    \centering
    \includegraphics[width=1.0\linewidth]{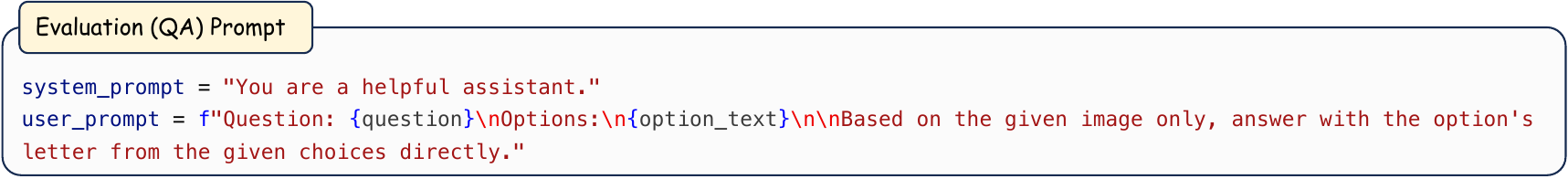}
    \vspace{-10pt}
    \caption{Prompt of QA evaluation.}
    \label{fig:eval_prompt}
    \vspace{-1.3em}
\end{figure*}

\noindent\textbf{Resolution Settings.}
For the image generation process, we adhere to the default resolution settings of each respective model to reflect their standard performance. Meanwhile, to ensure that the generated academic illustrations possess sufficient clarity and high-density detail for fine-grained logic, we enforced a minimum resolution threshold of 1024$\times$1024 pixels for all generated outputs.

\noindent\textbf{QA Evaluation Protocol.}
During the evaluation phase, we standardize the VQA process to guarantee fair comparisons and reliable reproducibility. A major limitation of the existing VLM-as-Judge paradigm is its reliance on complex, heavily engineered, and often brittle prompts to guide the judging model. In contrast, our VQA-based approach utilizes a single, straightforward QA template applied universally across all evaluated models, as illustrated in Fig.~\ref{fig:eval_prompt}. This unified template directly queries the VQA solver using our automatically generated and carefully human-checked multi-level question-answer pairs. By relying on objective visual queries rather than open-ended VLM judgments, we eliminate the need for intricate evaluation prompt tuning. This ensures that the resulting performance metrics strictly reflect the true logical consistency of the generated illustrations, rather than the efficacy of the evaluation prompt itself.

\begin{table*}[!t]
\centering
\caption{Extended evaluation of Test-Time Scaling strategies on AIBench. This table details the impact of the explicit reasoning phase ( \textit{e.g.}, text rewriting and SVG-based structural) priors across models with varying inherent capabilities, illustrating how the effectiveness of these enhancements heavily depends on the model's native comprehension and generation capacity.}
\label{tab:more_tts}
\setlength{\tabcolsep}{4pt} 
\renewcommand{\arraystretch}{1.2} 
\resizebox{0.94\textwidth}{!}{%
\begin{tabular}{l c c c c c >{\columncolor{cvprblue}}c}
\toprule
\textbf{Methods} & \textbf{Component} & \textbf{Topology} & \textbf{Phase} & \textbf{Semantics} & \textbf{Aesthetics} & \textbf{Overall} \\ \midrule
\multicolumn{6}{l}{\textcolor{gray}{\textit{{{Rewriting}}}}} \\
Wan2.6 \textit{w/o} Prompt Enhance & 35.29 & 42.70 & 64.66 & 79.64 & 42.58 & 52.97 \\
Gemini Rewritten Wan2.6 & 47.61 & 49.55 & 66.56 & 79.98 & \textbf{58.87} &  60.51 \\
Wan2.6~\cite{wan26} & \textbf{68.60} & \textbf{56.11} & \textbf{72.56} & \textbf{80.43} & 51.50 & \textbf{65.84} \\ \midrule
\multicolumn{6}{l}{\textcolor{gray}{\textit{{{AutoFigure Pipeline~\cite{zhu2026autofigure}}}}}} \\
SVG Code (Gemini-2.5-Flash~\cite{gemini2.5pro}) & 87.51 & 79.66 & 81.90 & 91.98 & 43.12 & 76.83 \\
Qwen-Image-2512~\cite{qwenimage} & 32.27 & 29.11 & 39.95 & 56.39 & 56.45 & 42.83 \\
\rowcolor[HTML]{dfdfdf} 
SVG Based Qwen-Image-2512 & 4.24 & 2.69 & 4.53 & 6.54 & 43.02 & 12.20 \\
Nano Banana~\cite{nano_banana} & 79.07 & 62.89 & 71.00 & 87.23 & 53.86 & 70.81 \\
\rowcolor[HTML]{dfdfdf} 
SVG Based Nano Banana & 76.90 & 66.99 & 75.88 & 87.30 & 53.09 & 72.03 \\
Nano Banana Pro~\cite{nano_banana}  & 87.80 & 74.81 & 82.67 & 88.54 & 55.04 & 77.77 \\
\rowcolor[HTML]{dfdfdf} 
SVG Prompted Nano Banana Pro & 87.57 & 76.50 & 78.91 & 92.14 & 55.05 & 78.03 \\ \bottomrule
\end{tabular}%
}
\end{table*}

\begin{figure*}[t]
    \centering
    \includegraphics[width=1.0\linewidth]{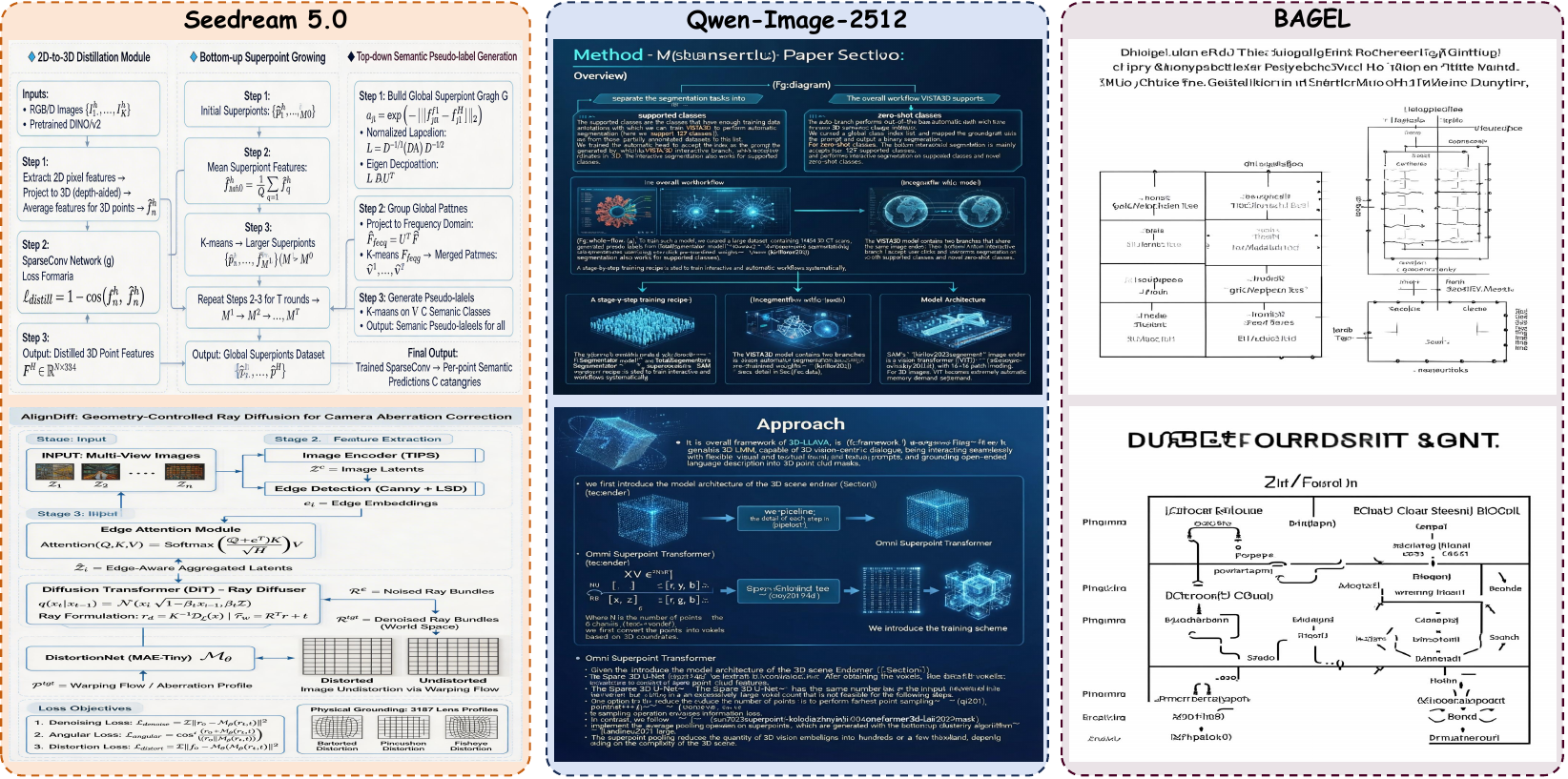}
    \vspace{-10pt}
    \caption{Qualitative comparison of academic illustrations generated by Seedream 5.0, Qwen-Image-2512, and BAGEL on AIBench.}
    \label{fig:trade-off between logical fidelity and aesthetics}
    \vspace{-1.3em}
\end{figure*}

\subsection{More Analysis on Test-Time Scaling}
To supplement the findings in the main paper, we provide a more granular analysis of the Test-Time Scaling (TTS) strategies, specifically exploring how different models react to text rewriting and intermediate structural priors (SVG code). The extended quantitative results are presented in Table~\ref{tab:more_tts}.

\noindent\textbf{The Efficacy of Rewriting Mechanisms.} 
While our main text demonstrates that explicit text rewriting significantly boosts the performance of models with weaker comprehension (\textit{e.g.}, Qwen-Image-2512), Table~\ref{tab:more_tts} reveals a different dynamic for state-of-the-art closed-source models. Advanced models like Wan2.6 are typically equipped with highly optimized internal prompt enhancement mechanisms. When we intentionally disable this native feature (\textit{Wan2.6 w/o Prompt Enhance}), the model's overall score experiences a severe drop from 65.84 to 52.97, particularly losing ground in logical dimensions such as Component and Topology. Furthermore, attempting to replace this native optimization with a generalized external LLM planner (\textit{Gemini Rewritten Wan2.6}) only partially recovers the performance (60.51), still underperforming the model's built-in mechanism. This confirms that explicit, external test-time rewriting is primarily crucial for open-source models lacking robust native comprehension, whereas top-tier models already possess near-optimal, deeply integrated text-processing pipelines.

\noindent\textbf{Model Capacity Dictates the Success of Structural Priors.} 
We also conduct a deeper investigation into the AutoFigure pipeline~\cite{zhu2026autofigure}, which utilizes LLM-generated SVG code as an intermediate structural prior. Our findings indicate that the effectiveness of this rigid, highly structured planning format is strictly dictated by the underlying generator's inherent capacity. 

For models with limited structural understanding, such as Qwen-Image-2512, injecting SVG code catastrophically disrupts the generation process. The model completely fails to interpret the abstract geometric instructions, causing the overall score to collapse from 42.83 to 12.20. Conversely, for models with solid foundational capabilities, the SVG prior serves as an effective spatial anchor; for instance, Nano Banana (used in the AutoFigure paper) shows a clear performance uplift (from 70.81 to 72.03) when guided by SVG. 

An interesting saturation effect occurs when scaling to exceptionally strong models like Nano Banana Pro. The performance gain becomes marginal (improving only slightly from 77.77 to 78.03). We attribute this to the fact that highly capable models already possess proficient end-to-end planning abilities. In such cases, forcing a hard intermediate translation step (Text $\rightarrow$ SVG Code $\rightarrow$ Image) may inadvertently act as an information bottleneck. The intermediate SVG generation is prone to omitting subtle methodological details or introducing structural hallucinations, thereby limiting the potential ceiling of a model that could otherwise directly render the text accurately.

\subsection{The Trade-off Between Logical Fidelity and Aesthetics}

As discussed in Section~\ref{subsec:main-results-aibench}, generating complex academic illustrations often involves a trade-off between logical fidelity and aesthetic quality. In this appendix, we provide additional examples to further demonstrate this phenomenon. As shown in Fig.~\ref{fig:trade-off between logical fidelity and aesthetics}, the closed-source model Seedream 5.0 produces illustrations with clearer logical organization, where individual steps and the overall progression are represented more accurately. However, this improvement in logical fidelity is accompanied by lower aesthetic quality. By contrast, Qwen-Image-2512 produces illustrations that are more visually appealing and exhibit a stronger technological style, leading to higher aesthetic scores. Yet its logical fidelity is comparatively weaker, making it less effective at clearly conveying the methodological content of the paper. In comparison, the unified model BAGEL shows inferior performance on both logical fidelity and aesthetic quality.

\begin{figure*}[!t]
    \centering
    \includegraphics[width=1.0\linewidth]{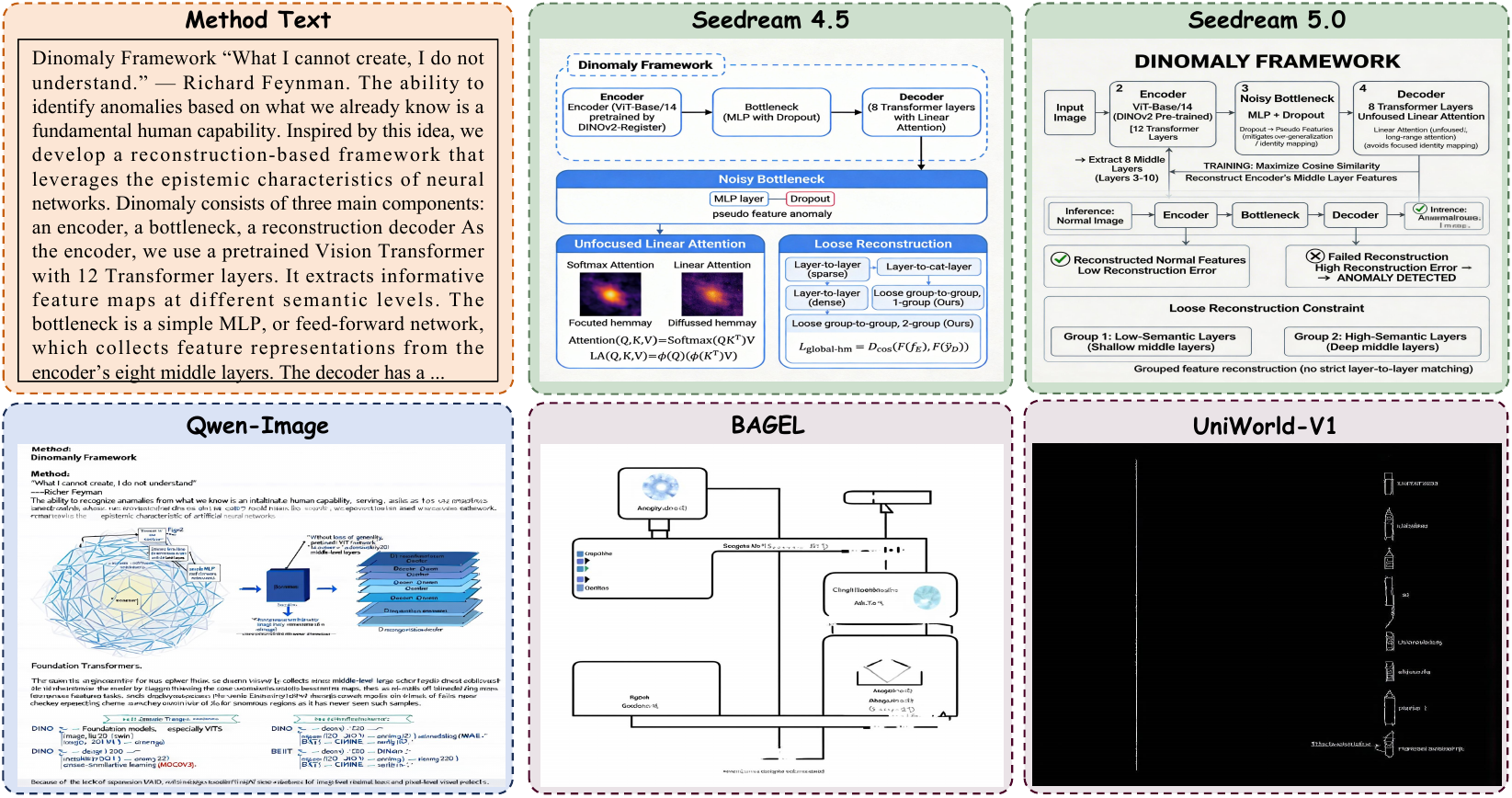}
    \vspace{-10pt}
    \caption{Qualitative comparison of academic illustrations generated by different models from the same method text on AIBench. The orange box shows the input method text, while the green, blue, and purple boxes present outputs from closed-source, open-source, and unified models, respectively.}
    \label{fig:case_method_text_models_1}
    \vspace{-1.3em}
\end{figure*}

\begin{figure*}[!t]
    \centering
    \includegraphics[width=1.0\linewidth]{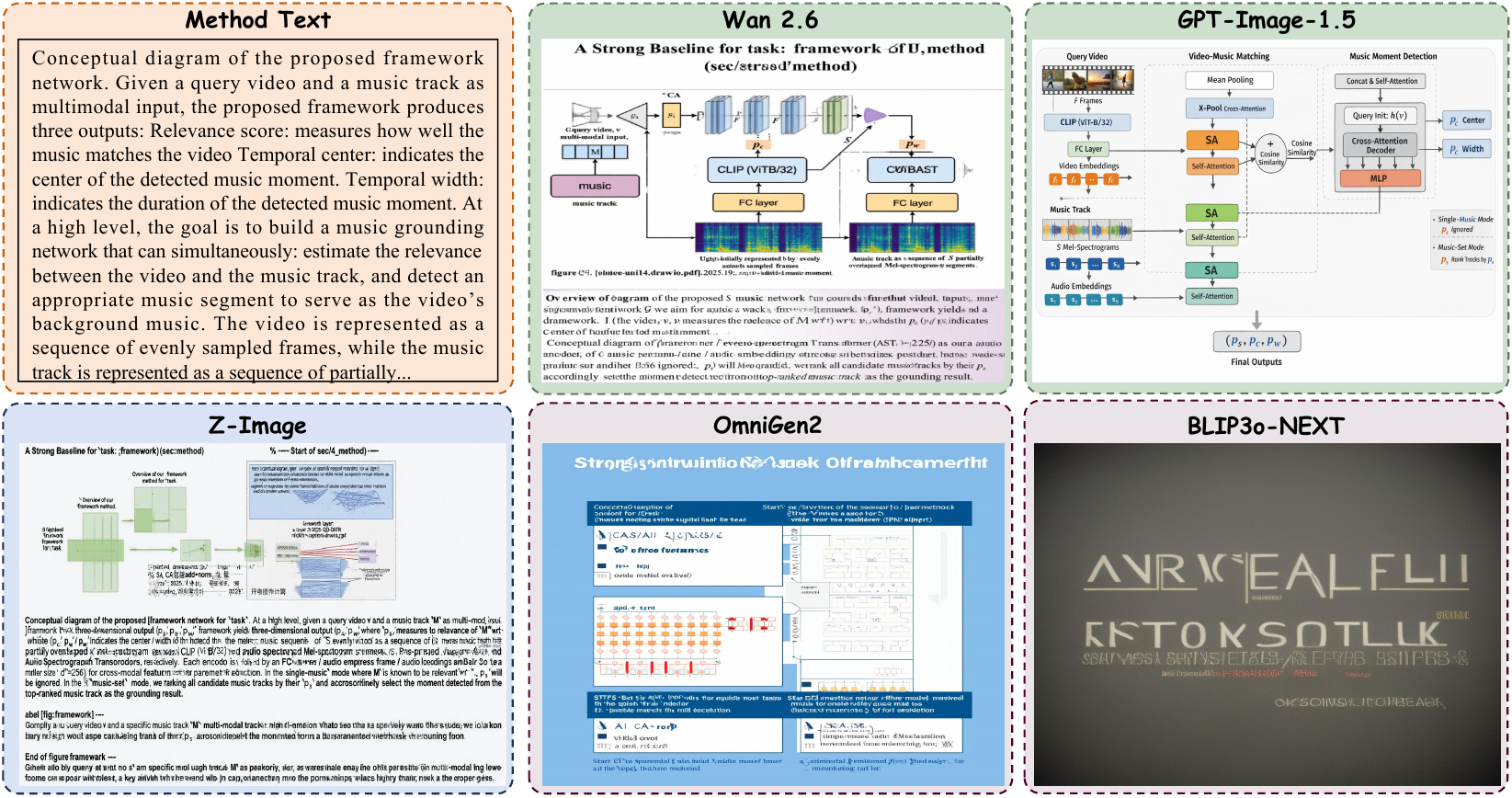}
    \vspace{-10pt}
    \caption{Qualitative comparison of academic illustrations generated by different models from the same method text on AIBench.}
    \label{fig:case_method_text_models_2}
\end{figure*}

\begin{figure*}[!t]
    \centering
    \includegraphics[width=1.0\linewidth]{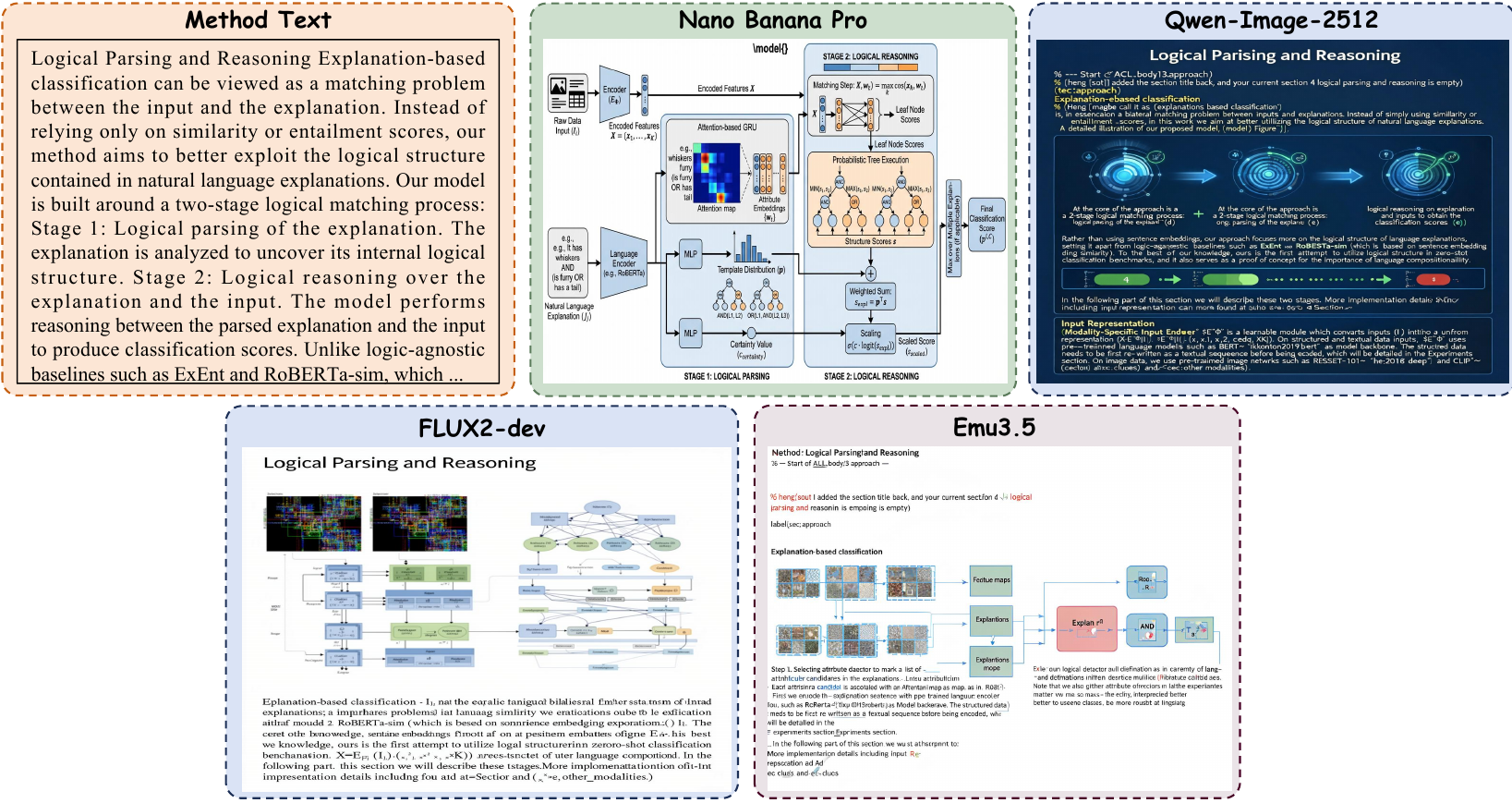}
    \vspace{-10pt}
    \caption{Qualitative comparison of academic illustrations generated by different models from the same method text on AIBench.}
    \label{fig:case_method_text_models_3}
    \vspace{-1.3em}
\end{figure*}

\begin{figure*}[!t]
    \centering
    \includegraphics[width=1.0\linewidth]{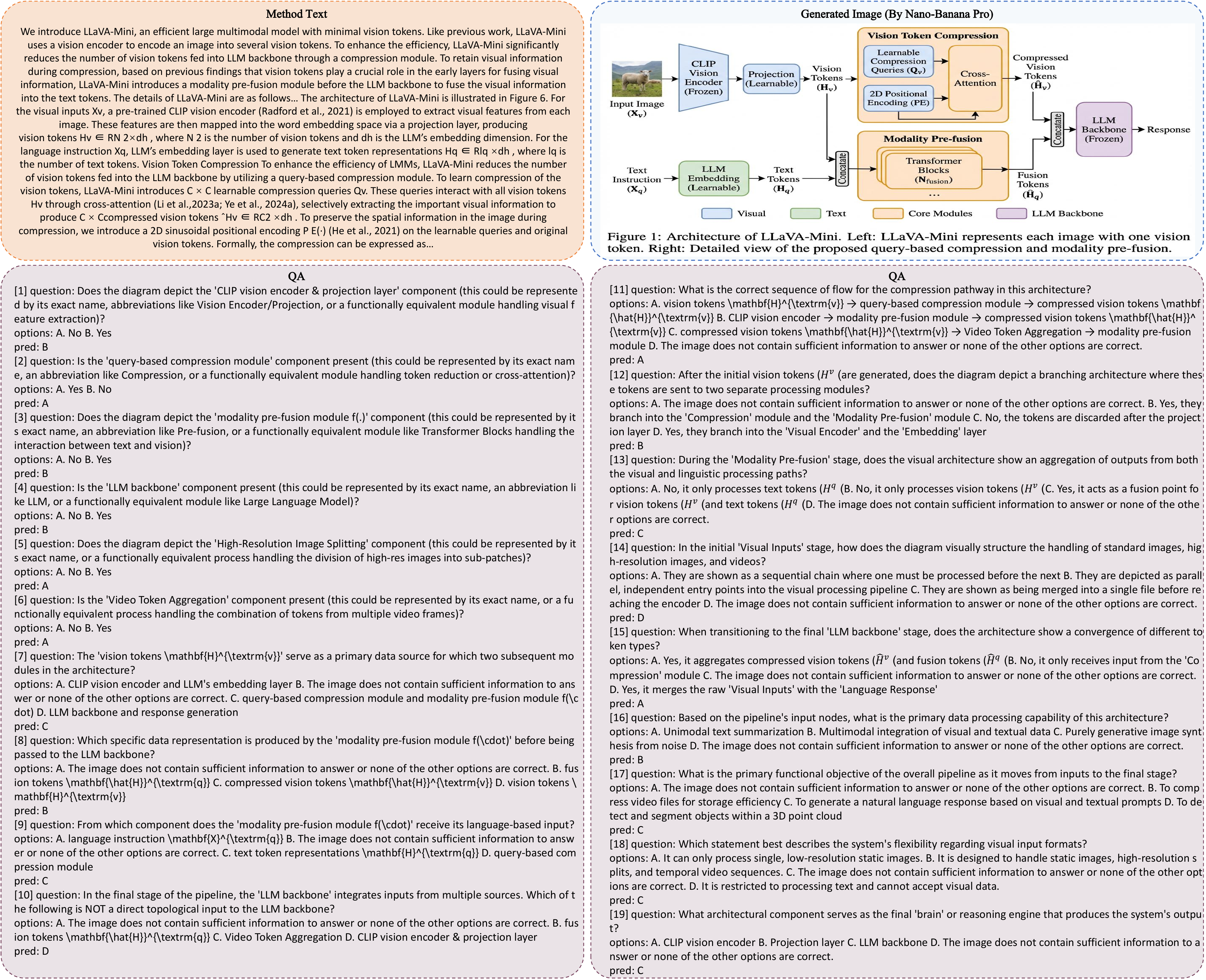}
    \vspace{-10pt}
    \caption{A complete AIBench case example. The case contains the \textbf{Method Text} (left), the \textbf{generated pipeline diagram} summarizing the method (middle), and an AIBench-curated \textbf{QA List} (right) that includes the evaluation \textbf{questions} and the corresponding \textbf{answers} grounded in the method text and pipeline.}
    \label{fig:whole_case}
    \vspace{-1.3em}
\end{figure*}

\subsection{Academic Illustrations Generated by Different Models on AIBench}

In this section, we present qualitative examples of academic illustrations generated by different models from the same method text on AIBench. As shown in Fig.~\ref{fig:case_method_text_models_1}, Fig.~\ref{fig:case_method_text_models_2}, and Fig.~\ref{fig:case_method_text_models_3}, closed-source models (green boxes) generally outperform open-source models (blue boxes) and unified models (purple boxes) in logical consistency and visual quality. Among the closed-source models, Nano Banana Pro produces illustrations that are strong in both structure and aesthetics, while GPT-Image-1.5 tends to generate cleaner and more concise figures, which is consistent with our quantitative findings. In contrast, open-source models still show substantial room for improvement in this task. Unified models are relatively limited in method text understanding and figure planning; for example, UniWorld-V1 and BLIP3o-NEXT sometimes produce semantically meaningless illustrations, which helps explain their poor performance on AIBench. We also observe that Emu3.5 often includes a large amount of method text in its generated academic illustrations, which may partially account for its relatively higher AIBench scores compared with other unified models.

\subsection{Illustration of a Case in AIBench}
To provide a concrete understanding of our dataset, we present a holistic case study in Figure~\ref{fig:whole_case}. Each entry in AIBench is a multi-modal ensemble designed to challenge the model's ability to cross-reference textual logic with visual structures. 
As illustrated, the case begins with the Method Text, which serves as the foundational description of the pipeline architecture. From this text, a corresponding Pipeline Image is generated. 
The core of our benchmark lies in the QA List. Unlike generic visual question-answering tasks, the questions in AIBench are \textit{meticulously constructed} to target deep structural reasoning. As shown in the figure, the questions require the model to identify specific components, trace the directional flow of information, and infer the relationship between different modules. For each question, we provide a precise answer derived from the generated image, ensuring a rigorous evaluation.

\section{Limitation and Future Work}

Although AIBench establishes a comprehensive and rigorous standard for evaluating academic illustration generation, it has a few limitations that pave the way for future exploration.
The current benchmark is curated from premier Artificial Intelligence conferences. While this ensures high-quality and dense methodological logic, it lacks representation from other scientific disciplines, such as biology, chemistry, and material sciences, which often rely on different diagrammatic conventions (\eg, molecular structures or complex experimental apparatuses).
In the future, we aim to expand the dataset to encompass a broader spectrum of academic fields to evaluate the cross-disciplinary generalization capabilities of generative models.

Furthermore, the insights derived from our experiments highlight critical pathways for the community to advance the methodological design of academic illustration generation. Future work should primarily focus on enhancing the reasoning capabilities of generative models to better handle long contexts and complex logic, alongside improving text rendering and structural generation abilities. Additionally, efforts should be made to elevate the aesthetic quality of generated outputs, extending this capability beyond academic illustrations to encompass a wider variety of diagrams, such as flowcharts. To effectively address these challenges, future research could explore architectural improvements, design more suitable test-time scaling paradigms tailored for academic illustration generation, and introduce reinforcement learning techniques to optimize both logical accuracy and aesthetic quality.

\section{Copyright and Licensing}
The source papers utilized to construct this benchmark are publicly accessible via arXiv under open-access licenses (\textit{e.g.}, CC BY 4.0). We strictly respect the intellectual property rights of the original authors. Our utilization of the methodology descriptions and diagrams is strictly limited to non-commercial, academic research. We rely on the principles of fair use and academic exceptions for text and data mining, ensuring that our benchmark serves as a transformative evaluation tool for the research community without infringing upon the original creators' rights.

\clearpage

\end{document}